\documentclass{article} 
\usepackage{iclr2027_conference,times}
\usepackage{booktabs}
\usepackage{tabularx}
\usepackage{array}
\usepackage{amssymb} 
\usepackage{wrapfig}
\usepackage{xcolor}
\usepackage{pifont}
\usepackage{tcolorbox}
\usepackage{graphicx}
\usepackage{soul}
\usepackage{multirow}
\usepackage{booktabs}
\usepackage{longtable}
\usepackage{array,ragged2e}
\usepackage{enumitem}
\usepackage{caption}
\usepackage[section]{placeins}
\usepackage{tcolorbox}

\tcbuselibrary{most,listings,breakable}

\definecolor{myframecolor}{RGB}{0,114,178}
\definecolor{pink-color}{RGB}{237,46,104}
\definecolor{dark-grey-color}{RGB}{79,91,102}
\definecolor{blue-color}{RGB}{0,114,178}

\newcommand*\colourcheck[1]{%
  \expandafter\newcommand\csname #1check\endcsname{\textcolor{#1}{\ding{51}}}%
}

\newcommand*\colourx[1]{%
  \expandafter\newcommand\csname #1x\endcsname{\textcolor{#1}{\ding{55}}}%
}

\colourcheck{green}
\colourx{red}

\newcommand{\promptsubsection}[1]{%
  \setlength{\parskip}{6pt}
  \noindent\textbf{#1:}
}

\newcommand{\paramnorm}[1]{%
  \textcolor{pink-color}{\small\texttt{\detokenize{#1}}}
}

\newtcolorbox[
  list inside=prompt,
  auto counter,
  number within=section
]{prompt}[1][]{
  colbacktitle=black!80,
  colframe=black!80,
  coltitle=white,
  fontupper=\footnotesize,
  boxsep=5pt,
  left=0pt,
  right=0pt,
  top=0pt,
  bottom=0pt,
  boxrule=1pt,
  enhanced,
  breakable,
  #1,
}

\usepackage{amsmath,amsfonts,bm}

\def\eqref#1{equation~\ref{#1}}

\def\1{\bm{1}}

\DeclareMathAlphabet{\mathsfit}{\encodingdefault}{\sfdefault}{m}{sl}
\SetMathAlphabet{\mathsfit}{bold}{\encodingdefault}{\sfdefault}{bx}{n}

\usepackage{hyperref}
\usepackage{url}

\usepackage[textsize=tiny]{todonotes}

\title{Quantifying Claim Smoothing by LLM\\ Research Agents}

\author{Tiasa Singha Roy \\
New York University\\
\texttt{ts5378@nyu.edu} \\
\And
Manya Wadhwa\\
New York University \\
\texttt{mw4141@nyu.edu} \\
\And
Greg Durrett \\
New York University \\
\texttt{gdurrett@nyu.edu}
}

\iclrfinalcopy 
\begin{document}

\maketitle

\begin{abstract}

Deep research, or any use of large language models (LLMs) to read a large document collection and produce insights, requires synthesizing information from different sources.  In this work, we find that LLMs produce long-form research reports that read as confident syntheses but which fail to accurately reflect the underlying documents. We call this failure \textit{smoothing}. In collapsing diverse and potentially conflicting sources into a single narrative, the model discards the evidential ``jaggedness'' of its corpus and asserts conclusions more decisive than the citations can support. We introduce a claim level scoring pipeline, \textsc{SmoothScore}, to quantify this effect. Our pipeline extracts key conclusive claims from model generated reports, then finds evidence relevant to each claim and clusters that evidence along high level aspects. Finally, each claim is scored with respect to the evidence to quantify the degree of smoothing done by the agent. Across 7 systems and 4 datasets, we find that agents consistently produce stronger claims than warranted\footnote{Code and data:  \url{https://github.com/tiasa2/SmoothScore}}.


\end{abstract}

\section{Introduction }

\begin{wrapfigure}{r}{0.35\textwidth}
\vspace{-0.15in}
    \centering
    \includegraphics[scale=0.4, trim= 1cm 1cm 36cm 0]{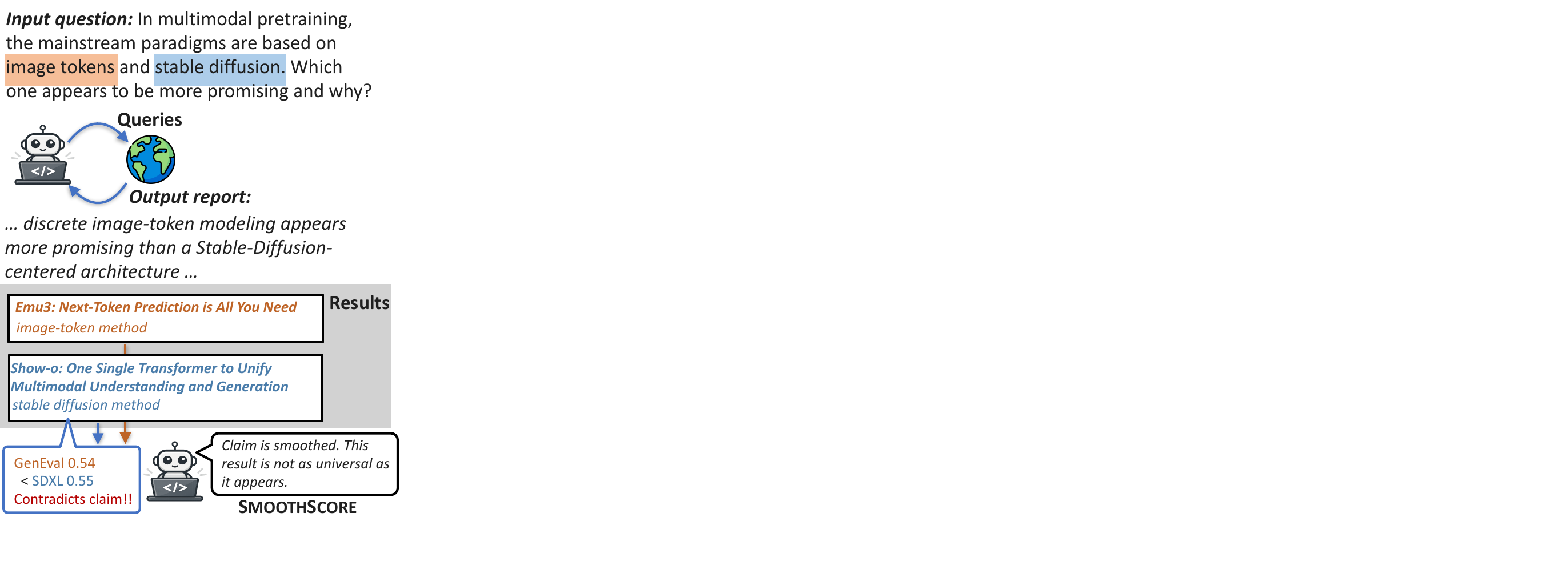}    \captionsetup{justification=raggedright,singlelinecheck=false}
    \caption{When synthesizing over many documents, agents can present a claim as well supported when the evidence is actually more mixed. \textsc{SmoothScore} flags these smoothed-over claims and quantifies the degree of smoothing.}
    \label{fig:intro}
    \vspace{-2em}
\end{wrapfigure}

Large Language Models (LLMs) are often deployed within systems that read across many sources of information and return long-form reports with citations~\citep{shao2026dr,xie2026quest,li2026webthinker,nakano2021webgpt}. These systems, either general-purpose agents paired with Internet access or special-purpose ``deep research'' agents, retrieve large volumes of documents and analyze and synthesize this information into coherent reports~\citep{du2026deepresearch,onweller2026cited}. The skills required to do this go beyond simple multi-document summarization ~\citep{fabbri2019multi}: agents must weigh evidence, integrate conflicting findings, and determine what the evidence warrants~\citep{narayanan2026deeptrace, rasheed2026fluent,zhu2026deep}. The resulting reports are typically fluent, well-organized, and densely cited, which makes them read as confident, authoritative syntheses. As a result, users increasingly treat these outputs as substitutes for reading the underlying literature, even when the synthesis misrepresents or overstates what the cited sources actually support~\citep{li2025human,huang2026deepfact}.

In this work, we study the prevalence of \textit{smoothing} in LLM generated reports. We define smoothing as the tendency of LLMs to synthesize claims that are more decisive than supported by the documents they cite. Such claims may, for example, fail to portray mixed evidence or overclaim a result to hold in all circumstances when it has only been shown more narrowly. By contrast, a faithful claim should hedge when the evidence is mixed and narrow its scope when the support is narrow.

To identify instances of this behavior, we propose \textsc{SmoothScore}, a framework for measuring whether the claims in a synthesized report are calibrated to the cited documents. As shown in Figure~\ref{fig:intro}, we  find that an agent's report can present a claim as well supported by some of its citations, even though other retrieved evidence shows the claim only holds under certain conditions. The bulk of prior work on claim verification and long-form factuality~\citep{min2023factscore,song2024veriscore, tang2024minicheck} can check whether individual claims are supported by their cited evidence, but breaks down for synthesized claims: these systems usually return binary judgments at the claim or subclaim level, but do not handle calibration and scoping across a large body of potentially conflicting evidence. 

To compute \textsc{SmoothScore}, we first decompose each report into its main conclusive claims and identify the documents the report cites. As shown in Figure~\ref{fig:Smoothscore_example}, for each claim, we run two agents over the report's cited documents: (a) \textit{Support Agent}: which surfaces evidence consistent with the claim, and (b) \textit{Skeptic Agent}: which actively searches for evidence that qualifies, limits, or contradicts the claim. Evidence from both these passes are clustered into aspects reflecting different dimensions of the claim, such as viewpoint, strength of argument, and certainty~\citep{kargupta2025truefalseretrievalaugmentedhierarchical}. 
Given the claim and its mapped aspect-evidence pairs, we score the claim's calibration on a 1–5 scale, reflecting how well its stated strength and scope match the evidence collected across both passes. 


Across four benchmarks and seven systems, we find that smoothing is pervasive. On average, for each system at least 25\% of generated claims received a \textsc{SmoothScore} $\geq 3$. For most systems over half of claims show at least material smoothing, and up to 35\% of Gemini Deep Research's claims show severe smoothing ($\geq 4$). We also find that limited interaction with the cited documents contributes to smoothing: Luna opens only 19\% of the pages it cites, and actively engaging with the full documents can help reduce smoothing. 

\textbf{Contributions} Our main contributions are: (a) We introduce \textsc{SmoothScore}, a framework for measuring whether the claims in a synthesized report are calibrated to the strength and scope of their cited evidence, rather than merely supported by it and (b) we find this issue to be widespread across frontier systems on 4 different benchmarks.


\section{Smoothing in LLM Research Agents} 
\label{sec:methodology}

\subsection{Defining Smoothing}


\paragraph{Setup and Definitions}
Given a prompt \(p\), a retrieval-augmented agent \(M\)  produces a
long-form response or report \(y\) with inline citations. We collect the full text of the cited documents in the report 
\(D = \{d_1,\ldots,d_m\}\).
Our goal is to measure if the synthesized conclusive claims in \(y\) faithfully
represent what \(D\) supports.


We call a claim \emph{smoothed} if it satisfies three criteria:(1) it reflects a conclusion drawn from aggregated evidence; (2) some of the evidence supports at least part of the conclusion; and (3) the conclusion as a whole is stronger or broader than the body of the evidence warrants.



This goal is closely related to hallucination detection
\citep{zha2023alignscore,min2023factscore,tang2024minicheck}; we view it as a special case of hallucination detection. However, typical methods for hallucination detection assume binary factuality judgments. Past work has looked at more nuanced judgments for narrative summarization \citep{subbiah2025spinningevaluatingsubjectivitynarrative} and opinion summarization \citep{bhaskar2023prompted}, but these methods are designed for simpler settings and do not handle the complexity of smoothing in deep research systems.

\begin{figure}[t]
    \centering
    \includegraphics[scale=0.33,trim=10cm 3cm 15cm 1cm]{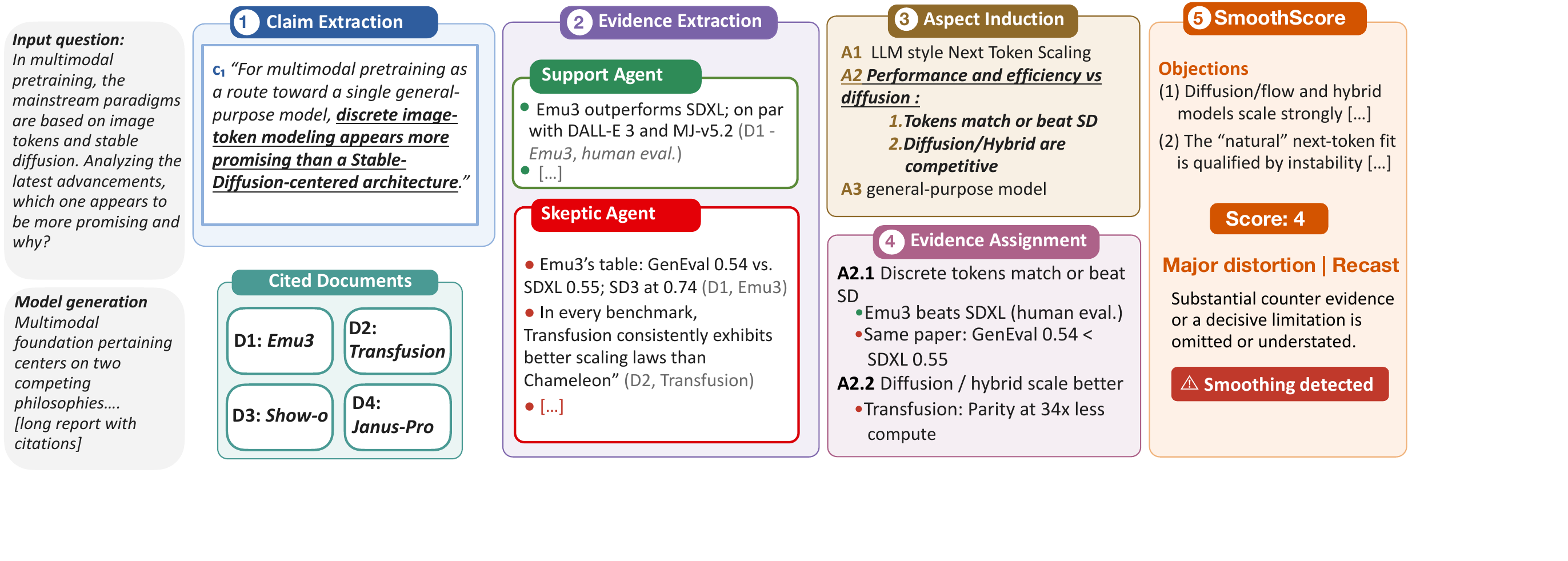}
    \caption{Overview of \textsc{SmoothScore}. Given a query and a system generated report, we (1) first extract conclusive claims. Then from the cited documents, (2) we get supporting and skeptic evidence relevant to the claim. (3)/(4) We then organize the evidence into \textit{aspects}. (5) The claim is then scored with respect to the gathered evidence to quantify the degree of missing information.}
    \label{fig:Smoothscore_example}
\end{figure}

Evaluating whether a claim is smoothed requires more than checking whether a report's statements are supported by individual citations. For instance, in Figure~\ref{fig:Smoothscore_example}, the extracted claim \textit{`discrete image-token modeling
appears more promising than a Stable-Diffusion-centered architecture'} is a conclusion made by the deep research system (satisfying condition (1) above). It is supported by the cited documents, $D_1$ Emu3 paper \citep{wang2024emu3nexttokenpredictionneed}) and $D_3$ (the Show-o paper \citep{xie2025showosingletransformerunify}) (satisfying condition (2)). However, we find that the same sources along with other documents, namely $D_2$ (the Transfusion paper \citep{zhou2024transfusionpredicttokendiffuse}), also contains results that show that the statement is only true under certain conditions present counter evidence (satisfying condition (3)).
Therefore, smoothing can occur even when a claim appears supported, and methods producing binary judgments from small evidence sets cannot comprehensively capture the distribution to identify it.

\paragraph{Identifying smoothing via aspects} We propose \textsc{SmoothScore} as  a way of measuring the degree of smoothing a conclusion shows. Figure~\ref{fig:Smoothscore_example} outlines the pipeline. At a high level, first, we extract such claims from a generated report that represent the key takeaways, then find evidence that supports and refutes those claims. However, it is not straightforward to combine this evidence into a final score: counting supporting documents vs.~refuting documents does not account for the nuances of \emph{how} a claim is supported or refuted.


Instead, for the claim $c$ in the report $y$ and evidence from a cited document $d \in D$, we identify a set of dimensions  $A(c,d)=\{a_1,\ldots,a_k\}$ that we call \emph{aspects}, following \citep{kargupta2025truefalseretrievalaugmentedhierarchical}. These aspects are consolidated across documents, to capture fine-grained ways in which a claim might be refuted. For instance, in Figure~\ref{fig:Smoothscore_example}, step 3, we are able to find \textit{`performance and efficiency of tokens vs diffusion models'}, under which we see some evidence state that \textit{`tokens match or beat diffusion'} but we also see evidence for \textit{`diffusion or hybrid set ups are competitive'}. As aspects are discovered per claim and tailored to the claim and its evidence,  we use them to score claim against the aspect-organized evidence to quantify its degree of smoothing.

\subsection{Measuring Smoothing with \textsc{SmoothScore}}


\paragraph{Step 1: Claim Extraction}  
To focus on the agents' ability to synthesize from aggregated evidence, we extract \emph{conclusion-bearing claims} (condition (1)). These are cases where report takes a stance or presents a takeaway. For example, in Figure~\ref{fig:Smoothscore_example}, \textit{`image-token modeling
appears more promising than a Stable-Diffusion-centered
architecture}' is a takeaway from the report which we extract as a claim.  


Using prompt, Appendix~\ref{sec:claim-generation}, we extract,  
\[
C = \operatorname{ExtractClaims}(y) = \{c_1,\ldots,c_n\}.
\]
Each claim preserves its direction, confidence, scope, and qualifications as stated in the report. 



\paragraph{Step 2: Evidence Extraction}  In Figure~\ref{fig:Smoothscore_example}, we see that a cited document, for example, $D1$ can carry information relevant to the claim that can both support or undermine it. To identify information relevant to the claim from the cited documents throughout the report, we extract evidence using a `support agent' and a `skeptic' agent. 

\begin{itemize}[leftmargin=*, itemsep=0pt, topsep=2pt]
    \item Support Agent (Appendix~\ref{sec:supporting_pass}): Given (claim, document), this agent extracts evidence that supports the assertions made by the claim. We also include stated concession as evidence when they confirm limitations or conditions present within the claim. 
\[
S(c) = \bigcup_{d \in D} \operatorname{Extract}_{+}(c,d)
\]
    
    \item Skeptic Agent (Appendix~\ref{sec:skeptic_pass}): Given (claim, document), this agent extracts evidence that contains any information against the claim. To capture the ``jaggedness'' we go beyond just finding counters to the assertions made by the claim. The agent also extracts evidence to support limitations and conditions stated in the claims. Additionally, we include evidence on alternate approaches and their success related to the claims where relevant. This is to provide context for scope.  
    \[
 R(c) = \bigcup_{d \in D} \operatorname{Extract}_{-}(c,d)
\]
\end{itemize}

For each agent, the evidence is composed of: a source quote from the document, aspect detailing the assertion, a note on how the evidence bears on the claim, and strength rating. 


\paragraph{Step 3: Aspect Induction}

Different cited documents can be relevant to different parts of a claim, so judging the claim against the raw documents forces the judge to compare evidence that is not directly comparable. If we look across the cited documents, however, the relevant information naturally clusters into topics that bear on the claim. For example, in Figure~\ref{fig:Smoothscore_example}, some documents discuss \textit{comparing the performance and efficiency of image tokens vs.\ diffusion approaches}, while others discuss \textit{fit with LLM-style next-token scaling}. We refer to these topics as \textit{aspects}~\citep{kargupta2025truefalseretrievalaugmentedhierarchical}.

To construct aspects, for each claim--document pair \((c_i,d_j)\in C\times D\) we generate a set of document-specific aspect proposals,
\[
A_{ij} = \operatorname{Propose}(c_i,d_j).
\]
Each proposal identifies a dimension discussed in \(d_j\) that bears on \(c_i\), and consists of a short name, a brief description, and a set of keywords. If \(d_j\) contains nothing relevant to \(c_i\), then \(A_{ij}=\varnothing\) (see Appendix~\ref{sec:Aspect_generation} for the prompt). Because proposals are generated independently for each document, different documents will often propose overlapping or near-duplicate aspects. For each claim, we therefore merge the proposals across documents into a single, deduplicated set of aspects, keeping track of which documents contributed to each one (see Appendix~\ref{sec:Aspect_merge}).

Organizing evidence by aspect places findings that address the same part of the claim side by side, regardless of which document they came from. This allows the judge to weigh evidence both supporting and opposing the claim on a like-for-like basis. For example, along a comparative dimension there may be supporting and countervailing evidence for either stance, as seen in aspect \textit{A2} of Figure~\ref{fig:Smoothscore_example}. 

\paragraph{Step 4: Evidence Assignment}

We assign every extracted evidence item in \(E(c_i)=S(c_i)\cup R(c_i)\) to the aspect it discusses, so that findings addressing the same part of the claim are grouped together and can be compared directly (prompt in Appendix~\ref{sec:Evidence_Assignment}). Assignment depends only on \emph{what} the evidence discusses, not on whether it supports or counters the claim, so an item's stance, source, and strength rating are left unchanged. For example, in Figure~\ref{fig:Smoothscore_example}, the support-agent finds that \textit{Emu3 beats SDXL in human evaluation} and the skeptic-agent finding that \textit{Emu3's own table shows a lower GenEval score than SDXL} are placed under the same aspect, even though they point in opposite directions.  



\paragraph{Step 5: Computing Smoothing}
Finally, a judge \(J\) produces a \textsc{SmoothScore} \(\sigma_i\) for each claim. The judge receives the input question \(q\), the claim \(c_i\), and all extracted evidence \(E_i\) organized under the claim's aspects \(T_i\), including evidence that is not tied to any single aspect:
\[
\sigma_i = J\bigl(q,c_i,T_i,E_i\bigr) \in \{1,\ldots,5\}.
\]
The judge assesses how faithfully the claim compresses the supplied evidence, focusing on whether it omits or understates assertions, conditions, or scope in a way that changes its central takeaway. Using the five-point rubric in ~\ref{sec:Smoothscore_judge} we score degree of smoothing (condition (3)). The criteria for receiving a score of 5 includes cases that are pure hallucination and don't match condition (2). However, we rarely observe cases of 5 (Section~\ref{sec:results}) and therefore view our system as primarily a method for analyzing smoothing rather than broader hallucination detection (even though it can identify such cases). Figure~\ref{fig:Smoothscore_example} and Table~\ref{tab:running-example} show examples of the generated outputs for each stage of the pipeline. 





\section{Experimental Setup}






\paragraph{Datasets.} We calculate \textsc{SmoothScore} on the following datasets: (1) \textbf{ResearcherBench} \citep{xu2025researcherbench}, (2) \textbf{ScholarQA-CS2 (Expert-Written)} \citep{hwang2026deep} (3) \textbf{DeepResearch Bench II} \citep{li2026deepresearchbenchiidiagnosing} (4) \textbf{ResearchRubrics} \citep{sharma2025researchrubricsbenchmarkpromptsrubrics}. These provide a range of tasks curated for research report generation. To ensure all queries require a degree of multi document synthesis we manually select queries. Details on this selection process is available in Appendix ~\ref{sec:dataset_curation} 



\paragraph{Systems.}
To check if overclaiming persists across architectures with varying design considerations, we evaluate the following 7 systems:
(1) \textbf{GPT Luna 5.6 + Websearch},
(2) \textbf{GPT Sol 5.6 + Websearch},
(3) \textbf{Claude Sonnet 5 + Websearch},
(4) \textbf{Claude Opus 5.5 + Websearch},
(5) \textbf{Gemini Deep Research},
(6) \textbf{DR Tulu} \citep{shao2026dr},
(7) and \textbf{QUEST} \citep{xie2026quest}. Together, these systems span several current approaches
to generating research reports, including agents augmented with Websearch, proprietary end-to-end deep-research systems, and open research agents with specialized retrieval and reasoning pipelines. Addtional details provided in Appendix~\ref{sec:metric_details}

\paragraph{Metrics} We report \textsc{SmoothScore} in two ways: (1) the mean \textsc{SmoothScore} across claims for each system and benchmark, and (2) the fraction of claims in each \textsc{SmoothScore} bin.  We believe the fraction of claims $\geq$3 indicates how often systems generate materially misleading claims.


\section{Results} \label{sec:results}





\subsection{\textsc{SmoothScore} Results}


\paragraph{Smoothing is widespread.} Table~\ref{tab:system-smoothing} shows that smoothing is pervasive across systems and datasets. Mean \textsc{SmoothScore} ranges from 1.66--1.95 for Luna and 1.82–1.93 for Sol to 3.05--3.27 for Gemini, with every other system falling between 2.20 and 2.88. The severity of smoothing is also reflected in the score distribution breakdown: smoothing of $\geq$3 is present on every system and dataset, with at least 22\% of claims affected in every case, but severe smoothing ($\geq$4) is where systems diverge sharply. Gemini smooths severely on 22--35\% of its claims, roughly 3–12 times the rate of other closed-source agents and 13-26× the rate of Sol.

\begin{table*}[t]
\centering
\scriptsize
\setlength{\tabcolsep}{3pt}
\caption{ We report the number of reports, claims for each of systems and benchmarks evaluated in this work. We also report the \textsc{SmoothScore} with $\geq$3 being materially smoothed and $\geq$4 being the subset of severely smoothed claims. Open source systems like, DR Tulu and QUEST produce a large \% of severely smoothed claims, while closed source agents vary with Luna and Sol systems only having 1-4\% claims $\geq 4$ against Gemini's 22-35\%.}
\label{tab:system-smoothing}
\resizebox{\textwidth}{!}{%
\begin{tabular}{llrr r rrrrr}
\toprule
& & \multicolumn{2}{c}{Sample} &
  \multicolumn{1}{c}{\textsc{SmoothScore}} &
  \multicolumn{5}{c}{\% of claims w/given \textsc{SmoothScore}} \\
\cmidrule(lr){3-4}
\cmidrule(lr){5-5}
\cmidrule(l){6-10}
Dataset & System & Reports & Claims &
Mean [95\% CI] & 1 & 2 & 3 & 4 & 5 \\
\midrule
\multirow{5}{*}{ResearcherBench}
 & Luna 5.6 & 65 & 443 & 1.95 [1.85, 2.06] & 37.0\% & 34.1\% & 25.7\% & 3.2\%  & 0.0\% \\
 & Sol 5.6  & 64 & 430 & 1.93 [1.81, 2.05] & 37.0\% & 35.6\% & 25.1\% & 2.3\%  & 0.0\% \\
 & Sonnet 5 & 65 & 435 & 2.52 [2.42, 2.61] & 13.8\% & 26.7\% & 53.3\% & 6.2\%  & 0.0\% \\
 & Opus 5.5 & 65 & 445 & 2.36 [2.28, 2.44] & 19.1\% & 29.4\% & 47.9\% & 3.6\%  & 0.0\% \\
 & QUEST    & 63 & 404 & 2.50 [2.35, 2.65] & 18.1\% & 26.7\% & 43.6\% & 10.1\% & 1.5\% \\
 & DR Tulu  & 65 & 408 & 2.57 [2.47, 2.67] & 12.0\% & 26.7\% & 54.2\% & 6.9\%  & 0.2\% \\
 & Gemini   & 64 & 434 & 3.27 [3.21, 3.34] & 1.6\%  & 5.1\%  & 58.1\% & 34.8\% & 0.5\% \\
\midrule
\multirow{5}{*}{CS2-Expert}
 & Luna 5.6 & 25 & 164 & 1.93 [1.75, 2.11] & 44.5\% & 23.8\% & 27.4\% & 3.0\%  & 1.2\% \\
 & Sol 5.6  & 25 & 165 & 1.82 [1.66, 1.98] & 43.6\% & 31.5\% & 23.6\% & 1.2\%  & 0.0\% \\
 & Sonnet 5 & 25 & 167 & 2.30 [2.12, 2.48] & 21.6\% & 32.3\% & 40.7\% & 5.4\%  & 0.0\% \\
 & Opus 5.5 & 25 & 188 & 2.20 [2.05, 2.33] & 28.7\% & 26.6\% & 42.0\% & 1.6\%  & 1.1\% \\
 & QUEST    & 25 & 156 & 2.67 [2.46, 2.87] & 17.3\% & 18.6\% & 47.4\% & 13.5\% & 3.2\% \\
 & DR Tulu  & 25 & 168 & 2.70 [2.53, 2.86] & 12.5\% & 19.0\% & 55.4\% & 12.5\% & 0.6\% \\
 & Gemini   & 24 & 175 & 3.18 [3.09, 3.28] & 4.0\%  & 5.7\%  & 59.4\% & 29.7\% & 1.1\% \\
\midrule
\multirow{7}{*}{DRB-II}
 & Luna 5.6 & 18 & 106 & 1.66 [1.46, 1.86] & 58.5\% & 18.9\% & 20.8\% & 1.9\%  & 0.0\% \\
 & Sol 5.6  & 18 & 111 & 1.83 [1.62, 2.04] & 45.0\% & 29.7\% & 23.4\% & 0.9\%  & 0.9\% \\
 & Sonnet 5 & 18 & 107 & 2.46 [2.24, 2.68] & 15.0\% & 30.8\% & 47.7\% & 6.5\%  & 0.0\% \\
 & Opus 5.5 & 18 & 106 & 2.39 [2.24, 2.53] & 21.7\% & 26.4\% & 43.4\% & 8.5\%  & 0.0\% \\
 & QUEST    & 16 &  95 & 2.88 [2.54, 3.25] & 9.5\%  & 22.1\% & 46.3\% & 14.7\% & 7.4\% \\
 & DR Tulu  & 18 & 102 & 2.49 [2.27, 2.71] & 19.6\% & 21.6\% & 50.0\% & 7.8\%  & 1.0\% \\
 & Gemini   & 18 & 120 & 3.05 [2.89, 3.23] & 2.5\%  & 12.5\% & 62.5\% & 22.5\% & 0.0\% \\
\midrule
\multirow{7}{*}{ResearchRubrics}
 & Luna 5.6 & 17 & 111 & 1.83 [1.57, 2.10] & 45.0\% & 31.5\% & 19.8\% & 2.7\%  & 0.9\% \\
 & Sol 5.6  & 17 & 108 & 1.93 [1.73, 2.12] & 36.1\% & 38.9\% & 22.2\% & 1.9\%  & 0.9\% \\
 & Sonnet 5 & 17 & 110 & 2.55 [2.37, 2.72] & 12.7\% & 29.1\% & 49.1\% & 8.2\%  & 0.9\% \\
 & Opus 5.5 & 17 & 100 & 2.42 [2.21, 2.61] & 17.0\% & 29.0\% & 49.0\% & 5.0\%  & 0.0\% \\
 & QUEST    & 17 & 104 & 2.88 [2.61, 3.16] & 12.5\% & 14.4\% & 50.0\% & 19.2\% & 3.8\% \\
 & DR Tulu  & 17 &  99 & 2.63 [2.38, 2.86] & 14.1\% & 24.2\% & 49.5\% & 9.1\%  & 3.0\% \\
 & Gemini   & 17 & 113 & 3.23 [3.05, 3.40] & 6.2\%  & 1.8\%  & 56.6\% & 33.6\% & 1.8\% \\
\bottomrule
\end{tabular}%
}
\end{table*}


\paragraph{Open-source agents produce severely smoothed claims} Systems follow similar patterns across datasets. The mean \textsc{SmoothScore} in open-source Deep Research systems like DR Tulu~\citep{shao2026dr} and QUEST~\citep{xie2026quest} remains comparable or slightly above that  Sonnet-5. However, at least $\sim$59\% and $\sim$55\% of their claims score $\geq 3$, and 7--23\% score $\geq 4$. From Table~\ref{tab:system-length-evidence}, QUEST draws from among the smallest evidence pools (roughly 10 documents and 40--55k words) so it has the least amount of evidence to compress. Despite this, while closed source systems only generate 1--2\% of claims that score $= 5$, QUEST generates up to $\sim$7\% and $\sim$4\% such claims on DeepResearchBench II and ResearchRubrics. Appendix~\ref{sec:open_source_systems} shows that QUEST often generates outputs that read as summarization rather than synthesis. The resulting claims preserve specifics faithfully while overstating scope enough to drive inflated conclusions.

\paragraph{Closed-source systems vary widely in smoothing severity.} Among closed-source agents with Websearch, we see that Sonnet 5 and Opus 5.5 produces almost 1.4--2.5$\times$ the fraction of claims that achieve a \textsc{SmoothScore} of 3--4 relative to Luna and Sol in Table~\ref{tab:system-smoothing}. These systems differ in the middle of the distribution rather than at its extremes and neither of the systems produce fundamentally misleading claims (Scored 5) at a meaningful rate, at most 1.1\% on any dataset. On the other hand, Luna and Sol generate at least 36\% claims that show no smoothing (score $= 1$), 1.5--2.7x more than any other system across datasets.

Across all systems, Gemini performs the worst, with 85--93\% of claims scoring $\geq 3$ and an average score of 3.05--3.27. Gemini also draws from significantly more documents (roughly 2–5x) than Luna or Sol with websearch, and produces the longest reports on every dataset (Table~\ref{tab:system-length-evidence}). This gives the agent a much larger surface to synthesize, and despite the added length, the report does not appear to compress this information more effectively. Appendix~\ref{sec:closed_source_systems} lists some examples of claims from  the generated reports. They contain language that turns a strong result into a universal or final conclusion.


\begin{wrapfigure}{r}{0.45\textwidth}
    \vspace{-1em}
    \centering
    \caption{Relationship between cited document count and claim smoothing. Across system and dataset pairs, we find that mean \textsc{SmoothScore} increases with the mean number of documents cited per report, but within a system this effect disappears (Table~\ref{tab:correlates}).} 
    \includegraphics[width=\linewidth]{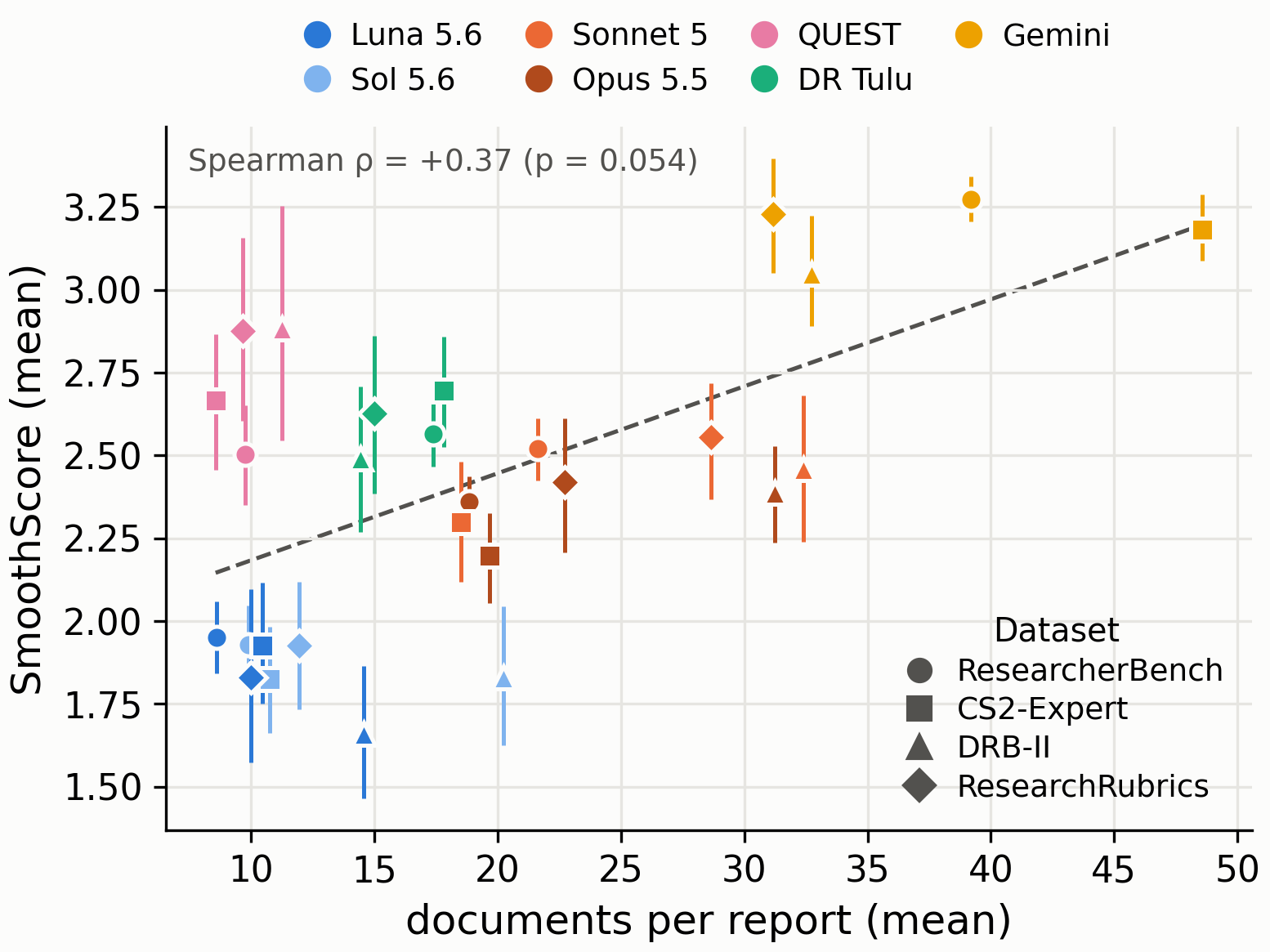}
    \label{fig:ladder_pooled}
    \vspace{-4em}
\end{wrapfigure}

\paragraph{Generation features show limited within-system associations with smoothing.} From Table~\ref{tab:system-length-evidence} we observe that the number of documents and report length vary across systems, which can influence the agents' ability to synthesize and therefore \textsc{SmoothScore}. We check if these features of an agent predict \textsc{SmoothScore} in Table~\ref{tab:correlates}. Across systems, systems that cite more documents ($\rho = 0.37$, $p = .056$) or write longer claims ($\rho = 0.43$, $p = .02$) smooth more, with Gemini at the extreme, but within a system both relationship are negligible ($\rho = 0.03$ and $0.02$; per-system values are small and mixed in sign). Only report length shows a within system association, with a small ($\rho = 0.12$) but positive correlation in all systems, though not significant in any one alone. Number of claims shows no such connection within or across systems (except a modest negative correlation for Gemini, $\rho = -0.23$). Overall, while these features show associations with smoothing, within-system  behavior suggests that other factors could also influence smoothing.


\subsection{Qualitative Observations}

\paragraph{Query framing influences \textsc{SmoothScore}.} We find that query framing influences how the model writes its claims. As seen in Appendix~\ref{sec:open_source_systems} and ~\ref{sec:closed_source_systems}, there are certain categories of questions where smoothing is more pronounced: (1) Comparative questions: if the query is comparative in nature, for example, \textit{`In multimodal pretraining, the paradigms are based on image tokens and stable diffusion, which appears more promising and why?'} or \textit{`Which paradigm between RL and distillation appears more promising and why?'}, it can presuppose a choice from the agent where to motivate or justify a choice the agent drops important qualifications, (2) Topic overview questions: queries that require an overview of a topic, for example, \textit{`Describe the Sunflower conjecture, progress towards a solution, and related problems'}, the agent can assemble multiple results a single conclusion or just generalize from a single finding to the whole topic.


\paragraph{Recurring Smoothing patterns across systems} 
Table~\ref{tab:smoothing-pattern-examples} shows three patterns that we observe across systems: (1) \textbf{Unsupported composition}:  Agents combine individual fragments of evidence into a stronger conclusion. Although each component may be supported, the claim presents its unsupported composition as a conclusive statement. (2) \textbf{Scope expansion}: A claim takes an established result for a particular setup or experimental condition and removes certain boundaries, so that a the scope for an empirical observation expands into a general conclusion. 
(3) \textbf{Premature Closure}: Claims often adopt conclusive language, using phrases like \textit{`strictly preserving,'} \textit{`the best approach,'} \textit{`definitively proves'} and \textit{`most optimal.'}.

\begin{table*}[t]
\small
\centering
\footnotesize
\caption{Examples of recurring claim-smoothing patterns. For each pattern, we show an illustrative claim, its \textsc{SmoothScore}, and a summary of the supporting evidence and limitations.}
\label{tab:smoothing-pattern-examples}
\begin{tabularx}{\textwidth}{
    >{\raggedright\arraybackslash}p{0.11\textwidth}
    >{\raggedright\arraybackslash}X
    >{\raggedright\arraybackslash}X}
\toprule
\textbf{Pattern} &
\textbf{Claim} &
\textbf{Summarized Evidence} \\
\midrule

Unsupported composition
&
\textit{The aggressive deployment of \textbf{duplication-aware token pruning},
\textbf{Modality-Aware Mixture of Experts}, and \textbf{adaptive halting} ensures that vast,
world-class parameter capacities \textbf{can be maintained without triggering
corresponding explosions in inference latency.}}
\newline\hfill\textbf{Score: 3}
&
Token pruning, sparse expert routing, and adaptive halting can each reduce
inference cost in particular settings. However, the evidence does not evaluate
their simultaneous deployment or show that their benefits compose. Each can
introduce routing, pruning, synchronization, or kernel overheads, so the
combined system does not guarantee bounded latency at frontier scale.
\\
\midrule

Scope expansion
&
\textit{Self-attention-based models are highly parallelizable and require
substantially fewer parameters, making them much more  \textbf{computationally
efficient}, \textbf{less prone to overfitting}, and \textbf{easier to fine-tune for
domain-specific tasks} \ldots{}}
\newline\hfill\textbf{Score: 4}
&
Some ViTs achieve strong performance with lower pre-training compute in
particular comparisons, but these advantages do not generalize across models
or regimes. ViTs often have comparable or larger parameter counts, global
attention has quadratic cost, and ViTs can be more prone to overfitting in
low-data settings without large-scale pre-training or careful regularization.
\\
\midrule

Premature closure
&
\textit{This innovation \textbf{fundamentally resolves the statistical heterogeneity
(Non-IID) problem} while strictly preserving the privacy boundaries established
by institutional firewalls.}
\newline\hfill\textbf{Score: 5}
&
Federated distillation can mitigate heterogeneity and reduce direct data
sharing, but it does not eliminate either problem. Performance continues to
degrade as heterogeneity increases, theoretical analyses retain residual bias,
and privacy attacks remain possible.
\\

\bottomrule
\end{tabularx}
\end{table*}

\section{Validating \textsc{SmoothScore}}
\label{sec:human_validation}

We validate \textsc{SmoothScore} by running a human study as well as checking the quality of evidence retrieved by our proposed pipeline.

\subsection{Human Validation}
\begin{wraptable}{l}{0.6\columnwidth}
\centering
\footnotesize
\setlength{\tabcolsep}{3pt}
\vspace{-1em}
\caption{Agreement on \textsc{SmoothScore} across 73 claims scored by two human annotators.}
\label{tab:agreement}
\begin{tabular}{lcccc}
\toprule
& \textbf{Exact} & \textbf{W/in 1} & $\boldsymbol{\rho}$ & $\boldsymbol{\alpha}$ \\
\midrule
Annotator 1 vs.\ Annotator 2 & 58\% & 96\% & 0.72 & 0.71 \\
Judge vs.\ Annotator 1       & 56\% & 93\% & 0.69 & 0.68 \\
\bottomrule
\end{tabular}
\vspace{-1em}
\end{wraptable}

To validate \textsc{SmoothScore}, two annotators independently scored the same 73 claims using the same rubric given to the LLM judge (Appendix~\ref{sec:Smoothscore_judge}), without access to the identity of the system that produced each claim. We find that agreement between the judge and Annotator~1 is in a similar range to inter-annotator agreement, Table~\ref{tab:agreement}. Disagreements are predominantly small: the judge is within one point of Annotator~1's score on 93\% of claims and differs by two or more points on only 7\%. Additionally, of the claims the judge scores $\geq 3$, 62\% are also scored $\geq 3$ by Annotator~1, indicating that most judge-flagged claims reflect genuine issues. Since the two annotators themselves agree at $\kappa = 0.62$ at this threshold, the judge disagreement is expected given the subjective
nature of the task.

\subsection{Relevance of evidence collected by \textsc{SmoothScore}}
\begin{table}[t]
\centering
\small
\setlength{\tabcolsep}{2pt}
\caption{Revising reports generated by Luna and Sonnet 5 on ResearcherBench, scored against gold rubrics by GPT-5. Simply providing evidence can increase rubric scores, matching or exceeding providing raw document while being a fraction of the size}
\label{tab:evidence-revision}

\begin{tabularx}{\columnwidth}{@{}>{\raggedright\arraybackslash}Xccc ccc@{}}
\toprule
& \multicolumn{3}{c}{\textbf{Luna}}
& \multicolumn{3}{c}{\textbf{Sonnet~5}} \\
\cmidrule(lr){2-4}\cmidrule(lr){5-7}
\textbf{Condition}
& \textbf{Context} & \textbf{Length} & $\boldsymbol{\Delta}$ ($p$)
& \textbf{Context} & \textbf{Length} & $\boldsymbol{\Delta}$ ($p$) \\
\midrule
WebSearch draft (score)
& -- & $1.0\times$ & 68.3
& -- & $1.0\times$ & 64.4 \\

Revise, no context
& -- & $1.6\times$ & $+0.4$ ($.22$)
& -- & $1.5\times$ & $+1.1$ ($.38$) \\

Revise + cited documents
& 117k & $2.2\times$ & $+4.1$ ($.004$)
& 324k & $2.4\times$ & $+9.7$ ($<.001$) \\

Revise + \textsc{SmoothScore} evidence
& 11k & $2.0\times$ & $+4.4$ ($<.001$)
& 65k & $2.9\times$ & $+12.0$ ($<.001$) \\
\bottomrule
\end{tabularx}
\end{table}


While human agreement validates the \textsc{SmoothScore} judge's final scores, we separately check whether the evidence \textsc{SmoothScore} surfaces (Steps 2--4) is useful. We consider an independent measure, using gold rubrics provided by ResearcherBench created for judging the overall quality of the generated report. We take Luna and Sonnet~5 WebSearch reports on the benchmark, and perform an additional revision step with no web search under three context settings: (1) no additional context, (2) providing all cited documents, and (3) the collected
\textsc{SmoothScore} evidence as context. We find that both context settings rise well above the no-context control ($+0.4$ and $+1.1$), with the extracted evidence improving rubric scores for both agents (Luna $+4.4$, Sonnet $+12.0$, both $p<.001$), matching or even exceeding revision with the full documents. The evidence achieves this with roughly one-fifth to one-tenth of the tokens of the full documents. While this does not test the quality of the evidence relative to the generated claims, it shows that the evidence captures information that matters to an external, expert-written evaluation independent of the judge's scores.

\section{Lack of engagement with document content}
\label{par:documnet_engagement}



The systems we evaluate generate reports with citations from the documents they retrieve. However, the agent's interaction with the cited material is hard to control. For Luna, the trace shows that the agent works almost entirely from search snippets truncated by the tool (median 585 characters), opening a cited page for only 19\% of the documents it cites and in just 20 of 65 reports. Sonnet~5 engages even less, opening only 2\% of the links it cites (Table~\ref{tab:document-access}, Appendix~\ref{app:document-engagement}). As agents rarely read the full documents, an agent may never see the qualifications and countervailing findings in the body of a source, which suggests that smoothing is partly shaped by what information the agent chooses to access. 

To test whether engaging with the documents reduces smoothing, we regenerate Luna and Sonnet~5 reports under varying setups. We vary (i) evidence access: the original WebSearch setting, or the original cited documents provided in context \textit{(cited docs in context)}, given in context with web search off; and (ii) instructions: the original prompt, or the same prompt plus calibration instructions. The instructions are identical across both evidence settings (prompts in Appendix~\ref{sec:docs_no_instructions},~\ref{sec:docs_with_instructions}). We score all reports with \textsc{SmoothScore} on ResearcherBench.

\begin{wraptable}{r}{0.55\columnwidth}
\centering\small
\setlength{\tabcolsep}{4pt}
\vspace{-1em}
\caption{Luna and sonnet-5 websearch system engagement with the cited documents on ResearcherBench. Luna's opens are marked by a header and Sonnet~5's opens are explicit web fetches returning the full page. Details in Appendix~\ref{app:document-engagement}.}
\label{tab:document-access}
\begin{tabular}{@{}lrr@{}}
\toprule
 & \textbf{Luna} & \textbf{Sonnet 5} \\
\midrule
Cited documents               & 496           & 1,458 \\
\quad Opened                  & 95 (19\%)     & 34 (2\%) \\
\quad Not opened              & 401 (81\%)    & 1,424 (98\%) \\
\midrule
Reports opening no page       & 45 / 65       & 44 / 65 \\
\bottomrule
\end{tabular}
\vspace{1em}
\end{wraptable}



\begin{table}[t!]
\centering\small
\setlength{\tabcolsep}{1.2mm}
\caption{Effect of document engagement and calibration instructions on smoothing
(ResearcherBench, 61 questions). \textbf{WebSearch}: original reports. \textbf{Docs}: the same agent given the full text of the documents its WebSearch report cited, with web search off.
\textbf{+ Instr.}: the same calibration instructions added to either prompt. Providing both shows the largest improvement for both setups.}
\label{tab:document-engagement}
\begin{tabular}{llcccccccc}
\toprule
& & & & & \multicolumn{5}{c}{\% of claims with \textsc{SmoothScore}} \\
\cmidrule(l){6-10}
Model & Condition & Length & Claims & Mean [95\% CI] & 1 & 2 & 3 & 4 & 5 \\
\midrule
\multirow{4}{*}{Luna}
 & WebSearch       & 1.0$\times$ & 416 & 1.95 [1.84, 2.07] & 37.3 & 33.4 & 26.0 & 3.4 & 0.0 \\
 & WebSearch + Instr.        & 1.0$\times$ & 397 & 1.88 [1.78, 1.98] & 40.3 & 34.0 & 23.2 & 2.5 & 0.0 \\
 & Docs            & 1.6$\times$ & 390 & 1.80 [1.70, 1.91] & 45.6 & 31.0 & 21.3 & 2.1 & 0.0 \\
 & Docs + Instr.   & 1.4$\times$ & 378 & 1.52 [1.44, 1.60] & 61.1 & 26.7 & 11.6 & 0.5 & 0.0 \\
\midrule
\multirow{4}{*}{Sonnet~5}
 & WebSearch       & 1.0$\times$ & 407 & 2.54 [2.44, 2.63] & 13.5 & 25.6 & 54.8 & 6.1 & 0.0 \\
 & WebSearch + Instr.        & 1.1$\times$ & 391 & 2.23 [2.12, 2.34] & 25.6 & 30.2 & 40.4 & 3.8 & 0.0 \\
 & Docs            & 1.2$\times$ & 394 & 2.38 [2.27, 2.49] & 18.3 & 31.0 & 45.4 & 5.3 & 0.0 \\
 & Docs + Instr.   & 1.3$\times$ & 367 & 1.78 [1.67, 1.88] & 49.9 & 25.1 & 22.9 & 1.9 & 0.3 \\
\bottomrule
\end{tabular}
\end{table}

We find that neither intervention is sufficient on its own (Table~\ref{tab:document-engagement}). For Luna, the instructions without the documents barely change smoothing (mean $\sigma$ 1.95 to 1.88), and the documents alone reduce it modestly (1.80). Sonnet~5 responds more to instructions alone (2.54 to 2.23) but little to the documents alone (2.38). Combined, however, documents and instructions reduce smoothing far more than either does separately: mean $\sigma$ falls to 1.52 for Luna and 1.78 for Sonnet~5, and the share of materially smoothed claims ($\sigma \geq 3$) drops from 29\% to 12\% and from 61\% to 25\%, removing more than half in both cases. Severe smoothing ($\sigma \geq 4$) nearly disappears (0.5\% and 2.2\%). Instructions to calibrate are thus most effective when the agent has the evidence needed to act on them.


Reports written from the full documents are longer (1.2 -1.6$\times$) but contain slightly fewer extracted claims, so the reduction does not come from shorter or emptier reports. We also observe this qualitatively, the claims do not necessarily change the takeaway. Instead they tend to drop conclusive language supporting the scope of the evidence in documents and provide the required limitations as needed. Table~\ref{tab:protection-example} shows that the central takeaway remains the same and smoothing tracks the added reasoning.

\begin{table}[t!]
\centering
\footnotesize
\renewcommand{\arraystretch}{1.1}
\caption{Claim improved under the docs+instruction case. Here the question pertains to how continual-learning systems can preserve previously learned knowledge while remaining adaptable to new tasks.The judge's verdict is shown on the WebSearch claim.}
\label{tab:protection-example}
\begin{tabular}{@{}p{0.22\linewidth}p{\dimexpr0.78\linewidth-2\tabcolsep\relax}@{}}
\toprule
\textbf{Condition} & \textbf{Claim} \\
\midrule
\textbf{WebSearch}\newline
SmoothScore: 4
& Protection should be soft and probabilistic, not permanently
frozen: an important parameter may eventually be revised if
sufficient contradictory evidence accumulates. \\
\addlinespace
\emph{Judge rationale}
& Requires recasting: the prescription excludes hard freezing
and non-probabilistic approaches despite evidence supporting
their effectiveness. \\
\midrule
\textbf{Docs + instructions}\newline
SmoothScore: 1
& Previously useful parameters or subnetworks should be protected
from destructive updates. This can be implemented through frozen
columns, masks, structural isolation, or soft importance-weighted
constraints [doc 2][doc 3][doc 5]. \\
\bottomrule
\end{tabular}
\end{table}

\section{Related Work}

\paragraph{Claim decomposition and aspect-based verification} A standard recipe for verifying long-form text is to decompose it into smaller checkable units and score each against a knowledge source ~\citep{min2023factscore, chen-etal-2024-complex, gunjal-durrett-2024-molecular}.  These methods generally return a binary decision per claim, which is a good granularity for factual precision, but it does not work for synthesized claims that can have varying degrees of smoothing. ConflictScore~\citep{liu2026conflictscore} is the closest to our setting, and it  extends this atomic claim recipe in a direction close to ours. They mark a claim as supported, contradicted or irrelevant per document retrieved, and aggregate these into claim-level conflict measures. However, a claim can be labeled \emph{supported} by every one of its cited documents and still smooth over information it omits. Our work instead follows \citet{kargupta2025truefalseretrievalaugmentedhierarchical} in organizing evidence along \emph{aspects} (e.g., viewpoint, strength, certainty), so that supporting and qualifying evidence on the feature of a  claim can be considered against each other.

\paragraph{Multi-Document Summarization} In science communication, prior work measures certainty in how findings are stated~\citep{pei-jurgens-2021-measuring} and models how information changes as findings move from papers to news and social media~\citep{wright-etal-2022-modeling}. For LLM summaries of scientific text, \citet{peters2025generalization} test ten widely used LLMs on about 4,900 summaries and find that most produce broader generalizations than the source texts, and that LLM summaries are nearly five times more likely than human-written ones to overgeneralize. In medicine, neural multi-document summaries of clinical trials are fluent but often factually inaccurate~\citep{wallace2021generating}, and LLM summaries of medical evidence can misstate the certainty of the evidence or overgeneralize by omitting attributes such as a disease subtype~\citep{tang2023evaluating}. These studies summarize a single document or a small set of abstracts. We extend the concern to long-form, multi-document research reports written by agents, and give a claim-level measure grounded in the documents a report cites.

\paragraph{Correctness evaluation for deep research systems} Agentic deep-research systems~\citep{du2026deepresearch, xu2025researcherbench, huang2026deepfact, li2026deepresearchbenchiidiagnosing, sharma2025researchrubricsbenchmarkpromptsrubrics} pair rubric-based insight coverage with citation grounded factuality checks. Several works audit the reliability of these systems. DeepTRACE~\citep{narayanan2026deeptrace} decomposes answers into statements and finds that deep-research agents remain one-sided on debate queries and include many statements unsupported by their listed sources. \citet{onweller2026cited} find that citations that resolve and are topically relevant often still fail a factual check against the cited content. \citet{zhu2026deep} show that a single misleading document can induce false conclusions in final reports. \citet{rasheed2026fluent} argue for claim-level auditability, including whether conflicts in the evidence are surfaced or suppressed. Across this literature, factuality is largely operationalized as citation validity: whether a claim is supported by some retrieved source. Our results show that this is not enough. Reports whose claims are individually supported can still present mixed or conditional evidence as a decisive conclusion, so assessing how faithfully the underlying evidence is \emph{presented} requires the claim-level, evidence-calibrated analysis we propose.

\section{Conclusion}

In this work, we studied the propensity of LLMs to \emph{smooth} their conclusions when generating reports based on sources. They frequently generate oversimplified statements or conclusions that aren't really backed up by the evidence cited. Our pipeline, \textsc{SmoothScore}, allows us to quantify this and establish its frequency across several state-of-the-art agents. Finally, we find that this models fail to read with their own cited documents, and suggest smoothing can improved by more thorough agents engaging wih their sources.

\section*{AI Use Statement}


Generative AI formed the backbone of the systems studied in this work. In addition, code written with generative AI was used to conduct and carry out the analyses here. However, the research hypotheses in this work and the design of the experiments were both human-originated. AI tools were used for assistance with revisions, but not for writing in a substantive way. The authors take responsibility for the final content of this work.

\section*{Acknowledgments} 

This work was supported by NSF CAREER Award IIS-2145280, NSF grant IIS-2433071, the NSF AI Institute for Foundations of Machine Learning (IFML), and the NSF under Cooperative Agreement 2421782 and the Simons Foundation grant MPS-AI-00010515 awarded to the NSF-Simons AI Institute for Cosmic Origins — CosmicAI. This work was also partially supported by the Sloan Foundation and grants from Amazon and Open Philanthropy. The research has been supported by a compute grant from NVIDIA and computing support from the Torch cluster at NYU. We also thank the members of the TAUR lab at NYU for their helpful discussions and feedback.

\bibliography{iclr2027_conference}

@inproceedings{min2023factscore,
  title={{FActScore: Fine-grained atomic evaluation of factual precision in long form text generation}},
  author={Min, Sewon and Krishna, Kalpesh and Lyu, Xinxi and Lewis, Mike and Yih, Wen-tau and Koh, Pang and Iyyer, Mohit and Zettlemoyer, Luke and Hajishirzi, Hannaneh},
  booktitle={Proceedings of the 2023 conference on empirical methods in natural language processing},
  pages={12076--12100},
  year={2023}
}

@inproceedings{chen-etal-2024-complex,
    title = {{Complex Claim Verification with Evidence Retrieved in the Wild}},
    author = "Chen, Jifan  and
      Kim, Grace  and
      Sriram, Aniruddh  and
      Durrett, Greg  and
      Choi, Eunsol",
    editor = "Duh, Kevin  and
      Gomez, Helena  and
      Bethard, Steven",
    booktitle = "Proceedings of the 2024 Conference of the North American Chapter of the Association for Computational Linguistics: Human Language Technologies (Volume 1: Long Papers)",
    month = jun,
    year = "2024",
    address = "Mexico City, Mexico",
    publisher = "Association for Computational Linguistics",
    url = "https://aclanthology.org/2024.naacl-long.196/",
    doi = "10.18653/v1/2024.naacl-long.196",
    pages = "3569--3587"
}

@inproceedings{gunjal-durrett-2024-molecular,
    title = {{"Molecular Facts: Desiderata for Decontextualization in {LLM} Fact Verification"}},
    author = "Gunjal, Anisha  and
      Durrett, Greg",
    editor = "Al-Onaizan, Yaser  and
      Bansal, Mohit  and
      Chen, Yun-Nung",
    booktitle = "Findings of the Association for Computational Linguistics: EMNLP 2024",
    month = nov,
    year = "2024",
    address = "Miami, Florida, USA",
    publisher = "Association for Computational Linguistics",
    url = "https://aclanthology.org/2024.findings-emnlp.215/",
    doi = "10.18653/v1/2024.findings-emnlp.215",
    pages = "3751--3768"
}

@article{liu2026conflictscore,
  title={{ConflictScore: Identifying and Measuring How Language Models Handle Conflicting Evidence}},
  author={Liu, Siyi and Halfaker, Aaron and Roth, Dan and Xia, Patrick},
  journal={arXiv preprint arXiv:2606.26437},
  year={2026}
}

@inproceedings{
du2026deepresearch,
title={{DeepResearch Bench: A Comprehensive Benchmark for Deep Research Agents}},
author={Mingxuan Du and Benfeng Xu and Chiwei Zhu and Licheng Zhang and Xiaorui Wang and Zhendong Mao},
booktitle={The Fourteenth International Conference on Learning Representations},
year={2026},
url={https://openreview.net/forum?id=hQ0K2Hhq7H}
}

@inproceedings{huang2026deepfact,
  title={{DeepFact: Co-Evolving Benchmarks and Agents for Deep Research Factuality}},
  author={Huang, Yukun and Ribeiro, Leonardo FR and Hardalov, Momchil and Dhingra, Bhuwan and Dreyer, Markus and Saligrama, Venkatesh},
  booktitle={Proceedings of the 64th Annual Meeting of the Association for Computational Linguistics (Volume 1: Long Papers)},
  pages={34356--34386},
  year={2026}
}

@misc{
xu2025researcherbench,
title={{ResearcherBench: Evaluating Deep {AI} Research Systems on the Frontiers of {AI} Research}},
author={Tianze Xu and Pengrui Lu and Lyumanshan Ye and Xiangkun Hu and Pengfei Liu},
year={2026},
url={https://openreview.net/forum?id=oj6A9hrNdL}
}

@article{hwang2026deep,
  title={{Deep Research, Shallow Evaluation: A Case Study in Meta-Evaluation for Long-Form QA Benchmarks}},
  author={Hwang, Jena D and Kishore, Varsha and Singh, Amanpreet and Haddad, Dany and Naik, Aakanksha and Hamada, Malachi and Bragg, Jonathan and D'Arcy, Mike and Weld, Daniel S and Wang, Lucy Lu and others},
  journal={arXiv preprint arXiv:2603.06942},
  year={2026}
}

@inproceedings{shao2026dr,
title={{{DR} Tulu: Reinforcement Learning with Evolving Rubrics for Deep Research}},
author={Rulin Shao and Akari Asai and Shannon Zejiang Shen and Hamish Ivison and Varsha Kishore and Jingming Zhuo and Xinran Zhao and Molly Park and Samuel G. Finlayson and David Sontag and Tyler Murray and Sewon Min and Pradeep Dasigi and Luca Soldaini and Faeze Brahman and Wen-tau Yih and Tongshuang Wu and Luke Zettlemoyer and Yoon Kim and Hannaneh Hajishirzi and Pang Wei Koh},
booktitle={Forty-third International Conference on Machine Learning},
year={2026},
url={https://openreview.net/forum?id=97NEP1pyS3}
}

@article{xie2026quest,
  title={{QUEST: Training Frontier Deep Research Agents with Fully Synthetic Tasks}},
  author={Xie, Jian and Lin, Tianhe and Wang, Zilu and Ning, Yuting and Yao, Yuekun and Xue, Tianci and Zhang, Zhehao and Li, Zhongyang and Zhang, Kai and Wu, Yufan and others},
  journal={arXiv preprint arXiv:2605.24218},
  year={2026}
}

@article{nakano2021webgpt,
  title={{WebGPT: Browser-assisted question-answering with human feedback}},
  author={Nakano, Reiichiro and Hilton, Jacob and Balaji, Suchir and Wu, Jeff and Ouyang, Long and Kim, Christina and Hesse, Christopher and Jain, Shantanu and Kosaraju, Vineet and Saunders, William and others},
  journal={arXiv preprint arXiv:2112.09332},
  year={2021}
}

@article{li2026webthinker,
  title={{Webthinker: Empowering large reasoning models with deep research capability}},
  author={Li, Xiaoxi and Jin, Jiajie and Dong, Guanting and Qian, Hongjin and Wu, Yongkang and Wen, Ji-Rong and Zhu, Yutao and Dou, Zhicheng},
  journal={Advances in Neural Information Processing Systems},
  volume={38},
  pages={120091--120131},
  year={2026}
}

@article{onweller2026cited,
  title={{Cited but Not Verified: Parsing and Evaluating Source Attribution in LLM Deep Research Agents}},
  author={Onweller, Hailey and Lumer, Elias and Huber, Austin and Ramchandani, Pia and Subbiah, Vamse Kumar and Feld, Corey},
  journal={arXiv preprint arXiv:2605.06635},
  year={2026}
}

@inproceedings{narayanan2026deeptrace,
  title={{DeepTRACE: Auditing Deep Research AI Systems for Tracking Reliability Across Citations and Evidence}},
  author={Narayanan Venkit, Pranav and Laban, Philippe and Zhou, Yilun and Huang, Kung-Hsiang and Mao, Yixin and Wu, Chien-Sheng},
  booktitle={International Conference on Learning Representations},
  volume={2026},
  pages={106041--106062},
  year={2026}
}

@article{rasheed2026fluent,
  title={{From Fluent to Verifiable: Claim-Level Auditability for Deep Research Agents}},
  author={Rasheed, Razeen A and Banerjee, Somnath and Mukherjee, Animesh and Hazra, Rima},
  journal={arXiv preprint arXiv:2602.13855},
  year={2026}
}

@article{li2025human,
  title={{Human Trust in AI Search: A Large-Scale Experiment}},
  author={Li, Haiwen and Aral, Sinan},
  journal={arXiv preprint arXiv:2504.06435},
  year={2025}
}

@article{zhu2026deep,
  title={{Is Deep Research Reliable? Misleading Knowledge Induces False Conclusions}},
  author={Zhu, Pengyu and Li, Lijun and Yang, Longju and Su, Sen and Shao, Jing},
  journal={arXiv preprint arXiv:2607.20891},
  year={2026}
}

@inproceedings{tang2024minicheck,
    title = "{M}ini{C}heck: Efficient Fact-Checking of {LLM}s on Grounding Documents",
    author = "Tang, Liyan  and
      Laban, Philippe  and
      Durrett, Greg",
    editor = "Al-Onaizan, Yaser  and
      Bansal, Mohit  and
      Chen, Yun-Nung",
    booktitle = "Proceedings of the 2024 Conference on Empirical Methods in Natural Language Processing",
    month = nov,
    year = "2024",
    address = "Miami, Florida, USA",
    publisher = "Association for Computational Linguistics",
    url = "https://aclanthology.org/2024.emnlp-main.499/",
    doi = "10.18653/v1/2024.emnlp-main.499",
    pages = "8818--8847"
}

@inproceedings{song2024veriscore,
    title = "{V}eri{S}core: Evaluating the factuality of verifiable claims in long-form text generation",
    author = "Song, Yixiao  and
      Kim, Yekyung  and
      Iyyer, Mohit",
    editor = "Al-Onaizan, Yaser  and
      Bansal, Mohit  and
      Chen, Yun-Nung",
    booktitle = "Findings of the Association for Computational Linguistics: EMNLP 2024",
    month = nov,
    year = "2024",
    address = "Miami, Florida, USA",
    publisher = "Association for Computational Linguistics",
    url = "https://aclanthology.org/2024.findings-emnlp.552/",
    doi = "10.18653/v1/2024.findings-emnlp.552",
    pages = "9447--9474"
}

@inproceedings{kargupta2025truefalseretrievalaugmentedhierarchical,
    title = {{Beyond True or False: Retrieval-Augmented Hierarchical Analysis of Nuanced Claims}},
    author = "Kargupta, Priyanka  and
      Tian, Runchu  and
      Han, Jiawei",
    editor = "Che, Wanxiang  and
      Nabende, Joyce  and
      Shutova, Ekaterina  and
      Pilehvar, Mohammad Taher",
    booktitle = "Proceedings of the 63rd Annual Meeting of the Association for Computational Linguistics (Volume 1: Long Papers)",
    month = jul,
    year = "2025",
    address = "Vienna, Austria",
    publisher = "Association for Computational Linguistics",
    url = "https://aclanthology.org/2025.acl-long.1434/",
    doi = "10.18653/v1/2025.acl-long.1434",
    pages = "29664--29679",
    ISBN = "979-8-89176-251-0"
}

@inproceedings{fabbri2019multi,
    title = "Multi-News: A Large-Scale Multi-Document Summarization Dataset and Abstractive Hierarchical Model",
    author = "Fabbri, Alexander  and
      Li, Irene  and
      She, Tianwei  and
      Li, Suyi  and
      Radev, Dragomir",
    editor = "Korhonen, Anna  and
      Traum, David  and
      M{\`a}rquez, Llu{\'i}s",
    booktitle = "Proceedings of the 57th Annual Meeting of the Association for Computational Linguistics",
    month = jul,
    year = "2019",
    address = "Florence, Italy",
    publisher = "Association for Computational Linguistics",
    url = "https://aclanthology.org/P19-1102/",
    doi = "10.18653/v1/P19-1102",
    pages = "1074--1084"
}

@misc{li2026deepresearchbenchiidiagnosing,
      title={{DeepResearch Bench II: Diagnosing Deep Research Agents via Rubrics from Expert Reports}}, 
      author={Ruizhe Li and Mingxuan Du and Benfeng Xu and Chiwei Zhu and Xiaorui Wang and Zhendong Mao},
      year={2026},
      eprint={2601.08536},
      archivePrefix={arXiv},
      primaryClass={cs.CL},
      url={https://arxiv.org/abs/2601.08536}, 
}

@misc{sharma2025researchrubricsbenchmarkpromptsrubrics,
      title={{ResearchRubrics: A Benchmark of Prompts and Rubrics For Evaluating Deep Research Agents}}, 
      author={Manasi Sharma and Chen Bo Calvin Zhang and Chaithanya Bandi and Clinton Wang and Ankit Aich and Huy Nghiem and Tahseen Rabbani and Ye Htet and Brian Jang and Sumana Basu and Aishwarya Balwani and Denis Peskoff and Marcos Ayestaran and Sean M. Hendryx and Brad Kenstler and Bing Liu},
      year={2025},
      eprint={2511.07685},
      archivePrefix={arXiv},
      primaryClass={cs.AI},
      url={https://arxiv.org/abs/2511.07685}, 
}

@misc{xie2025showosingletransformerunify,
      title={{Show-o: One Single Transformer to Unify Multimodal Understanding and Generation}}, 
      author={Jinheng Xie and Weijia Mao and Zechen Bai and David Junhao Zhang and Weihao Wang and Kevin Qinghong Lin and Yuchao Gu and Zhijie Chen and Zhenheng Yang and Mike Zheng Shou},
      year={2025},
      eprint={2408.12528},
      archivePrefix={arXiv},
      primaryClass={cs.CV},
      url={https://arxiv.org/abs/2408.12528}, 
}

@article{wang2024emu3nexttokenpredictionneed,
  publtype={informal},
  author={Xinlong Wang and Xiaosong Zhang and Zhengxiong Luo and Quan Sun and Yufeng Cui and Jinsheng Wang and Fan Zhang and Yueze Wang and Zhen Li and Qiying Yu and Yingli Zhao and Yulong Ao and Xuebin Min and Tao Li and Boya Wu and Bo Zhao and Bowen Zhang and Liangdong Wang and Guang Liu and Zheqi He and Xi Yang and Jingjing Liu and Yonghua Lin and Tiejun Huang and Zhongyuan Wang},
  title={{Emu3: Next-Token Prediction is All You Need}},
  year={2024},
  cdate={1704067200000},
  journal={CoRR},
  volume={abs/2409.18869},
  url={https://doi.org/10.48550/arXiv.2409.18869}
}

@misc{zhou2024transfusionpredicttokendiffuse,
      title={Transfusion: Predict the Next Token and Diffuse Images with One Multi-Modal Model}, 
      author={Chunting Zhou and Lili Yu and Arun Babu and Kushal Tirumala and Michihiro Yasunaga and Leonid Shamis and Jacob Kahn and Xuezhe Ma and Luke Zettlemoyer and Omer Levy},
      year={2024},
      eprint={2408.11039},
      archivePrefix={arXiv},
      primaryClass={cs.AI},
      url={https://arxiv.org/abs/2408.11039}, 
}

@inproceedings{pei-jurgens-2021-measuring,
    title = "Measuring Sentence-Level and Aspect-Level (Un)certainty in Science Communications",
    author = "Pei, Jiaxin  and
      Jurgens, David",
    editor = "Moens, Marie-Francine  and
      Huang, Xuanjing  and
      Specia, Lucia  and
      Yih, Scott Wen-tau",
    booktitle = "Proceedings of the 2021 Conference on Empirical Methods in Natural Language Processing",
    month = nov,
    year = "2021",
    address = "Online and Punta Cana, Dominican Republic",
    publisher = "Association for Computational Linguistics",
    url = "https://aclanthology.org/2021.emnlp-main.784/",
    doi = "10.18653/v1/2021.emnlp-main.784",
    pages = "9959--10011"
}

@inproceedings{wright-etal-2022-modeling,
    title = "Modeling Information Change in Science Communication with Semantically Matched Paraphrases",
    author = "Wright, Dustin  and
      Pei, Jiaxin  and
      Jurgens, David  and
      Augenstein, Isabelle",
    editor = "Goldberg, Yoav  and
      Kozareva, Zornitsa  and
      Zhang, Yue",
    booktitle = "Proceedings of the 2022 Conference on Empirical Methods in Natural Language Processing",
    month = dec,
    year = "2022",
    address = "Abu Dhabi, United Arab Emirates",
    publisher = "Association for Computational Linguistics",
    url = "https://aclanthology.org/2022.emnlp-main.117/",
    doi = "10.18653/v1/2022.emnlp-main.117",
    pages = "1783--1807"
}

@article{peters2025generalization,
  title={{Generalization Bias in Large Language Model Summarization of Scientific Research}},
  author={Peters, Uwe and Chin-Yee, Benjamin},
  journal={Royal Society Open Science},
  volume={12},
  number={4},
  pages={241776},
  year={2025}
}

@article{wallace2021generating,
  title={{Generating (Factual?) Narrative Summaries of RCTs: Experiments with Neural Multi-Document Summarization}},
  author={Wallace, Byron C and Saha, Sayantan and Soboczenski, Frank and Marshall, Iain J},
  journal={AMIA Summits on Translational Science Proceedings},
  volume={2021},
  pages={605},
  year={2021}
}

@article{tang2023evaluating,
  title={{Evaluating Large Language Models on Medical Evidence Summarization}},
  author={Tang, Liyan and Sun, Zhaoyi and Idnay, Betina and Nestor, Jordan G and Soroush, Ali and Elias, Pierre A and Xu, Ziyang and Ding, Ying and Durrett, Greg and Rousseau, Justin F and others},
  journal={NPJ digital medicine},
  volume={6},
  number={1},
  pages={158},
  year={2023},
  publisher={Nature Publishing Group UK London}
}

@misc{openai2026websearch,
  author       = {{OpenAI}},
  title        = {Web search},
  howpublished = {\url{https://developers.openai.com/api/docs/guides/tools-web-search}},
  year         = {2026},
  note         = {OpenAI API documentation. Accessed October 2026}
}

@misc{anthropic2026websearch,
  author       = {{Anthropic}},
  title        = {Web search tool},
  howpublished = {\url{https://platform.claude.com/docs/en/agents-and-tools/tool-use/web-search-tool}},
  year         = {2026},
  note         = {Claude API documentation. Accessed October 2026}
}

@misc{anthropic2026webfetch,
  author       = {{Anthropic}},
  title        = {Web fetch tool},
  howpublished = {\url{https://platform.claude.com/docs/en/agents-and-tools/tool-use/web-fetch-tool}},
  year         = {2026},
  note         = {Claude API documentation. Accessed October 2026}
}

@inproceedings{subbiah2025spinningevaluatingsubjectivitynarrative,
    title = {{Is the Top Still Spinning? Evaluating Subjectivity in Narrative Understanding}},
    author = "Subbiah, Melanie  and
      Mishra, Akankshya  and
      Kim, Grace  and
      Tang, Liyan  and
      Durrett, Greg  and
      McKeown, Kathleen",
    editor = "Christodoulopoulos, Christos  and
      Chakraborty, Tanmoy  and
      Rose, Carolyn  and
      Peng, Violet",
    booktitle = "Proceedings of the 2025 Conference on Empirical Methods in Natural Language Processing",
    month = nov,
    year = "2025",
    address = "Suzhou, China",
    publisher = "Association for Computational Linguistics",
    url = "https://aclanthology.org/2025.emnlp-main.10/",
    doi = "10.18653/v1/2025.emnlp-main.10",
    pages = "185--203",
    ISBN = "979-8-89176-332-6",
}

@inproceedings{bhaskar2023prompted,
  title={Prompted opinion summarization with GPT-3.5},
  author={Bhaskar, Adithya and Fabbri, Alex and Durrett, Greg},
  booktitle={Findings of the association for computational linguistics: Acl 2023},
  pages={9282--9300},
  year={2023}
}

@inproceedings{zha2023alignscore,
  title={AlignScore: Evaluating factual consistency with a unified alignment function},
  author={Zha, Yuheng and Yang, Yichi and Li, Ruichen and Hu, Zhiting},
  booktitle={Proceedings of the 61st annual meeting of the association for computational linguistics (volume 1: Long papers)},
  pages={11328--11348},
  year={2023}
}
\bibliographystyle{iclr2027_conference}

\appendix

\section{SmoothScore}

\subsection{Method Details}

\paragraph{Aspect Induction} As mentioned in 
Section~\ref{sec:methodology}, the third step in our pipeline is Aspect Induction. Since aspect proposals are generated separately for each claim--document pair, the same underlying topic is often proposed by several documents under different names or with different levels of specificity. Merging is needed so that evidence on the same topic is grouped together, rather than being split across near-duplicate aspects.

For each claim \(c_i\), we consolidate the document-level proposals \(\{A_{ij}\}_{j=1}^{m}\) into a hierarchy
\[
T_i=(V_i,L_i)=\operatorname{Merge}\left(\{A_{ij}\}_{j=1}^{m}\right),
\]
where \(V_i\) is the set of aspect and subaspect nodes and \(L_i\) is the set of parent--child links between them. The merge step (i) combines semantically overlapping proposals across documents into a single node, and (ii) introduces subaspects when one aspect covers distinct but related findings. Each node retains the set of documents that proposed it, so that every piece of evidence can be traced back to its source.

\paragraph{Pipeline Models}
Table~\ref{tab:pipeline-models} shows the models used for every stage of the pipeline for  generating \textsc{SmoothScore}. All pipeline stages use GPT-5-mini except  \textsc{SmoothScore} judge (Step 5), which use GPT-5; every call uses \texttt{reasoning\_effort=medium} with default sampling.

\subsection{Dataset Curation}
\label{sec:dataset_curation}
We use all 65 ResearcherBench questions and the complete 25-question expert-written ScholarQA-CS2 set. These by construction contained open ended questions that required multi document engagement. For DeepResearch Bench II and ResearchRubrics: 
 \begin{itemize}
  \item \textbf{Candidate pool.} We take the 66 English DeepResearch Bench II tasks and all
  101 ResearchRubrics tasks.
  \item \textbf{Annotation.} We label each query on synthesis (\textsc{High} / \textsc{Med} /
  \textsc{Low}), meaning whether a good answer must synthesize a body of literature. We also
  label openness of question type (\textsc{High} / \textsc{Med} / \textsc{Low}), meaning whether the question requires one definitive correct response.
  \item \textbf{Filter.} We keep queries with synthesis $=$ \textsc{High} and Question Type $=$ \textsc{High} (Table~\ref{tab:query-selection}).
\end{itemize}

\begin{table}[t]
\centering
\small
\setlength{\tabcolsep}{6pt}
\caption{Models used at each stage of the \textsc{SmoothScore} pipeline. All stages use GPT-5-mini except the final judge, which uses GPT-5. Prompts for each stage are given in the referenced appendix sections.}
\label{tab:pipeline-models}
\begin{tabular}{llll}
\toprule
Step & Stage & Model & Prompt \\
\midrule
1 & Claim extraction            & GPT-5-mini & App.~\ref{sec:claim-generation} \\
2 & Evidence extraction: support agent & GPT-5-mini & App.~\ref{sec:supporting_pass} \\
2 & Evidence extraction: skeptic agent & GPT-5-mini & App.~\ref{sec:skeptic_pass} \\
3 & Aspect induction: per-document proposals & GPT-5-mini & App.~\ref{sec:Aspect_generation} \\
3 & Aspect induction: cross-document merge   & GPT-5-mini & App.~\ref{sec:Aspect_merge} \\
4 & Evidence assignment         & GPT-5-mini & App.~\ref{sec:Evidence_Assignment} \\
5 & \textsc{SmoothScore} judge  & GPT-5      & App.~\ref{sec:Smoothscore_judge} \\
\bottomrule
\end{tabular}
\end{table}

\begin{table}[h]
\centering
\small
\caption{Query selection for DeepResearch Bench II and ResearchRubrics.}
\begin{tabular}{lrr}
\toprule
\textbf{Benchmark} & \textbf{Candidates} & \textbf{Kept} \\
\midrule
DeepResearch Bench II (English) & 66  & 18  \\
ResearchRubrics                 & 101 & 17  \\
\bottomrule
\end{tabular}
\label{tab:query-selection}
\end{table}

\subsection{Models}
\label{sec:metric_details}
Table~\ref{tab:system-ids} contains model IDs of the evaluated systems. We evaluate 7 systems consisting of open sourced and close source agents including frontier models.

\begin{table*}[h]
\centering
\small
\caption{Model identifiers and access method for each evaluated system.}
\begin{tabular}{lll}
\toprule
\textbf{System} & \textbf{Model ID} & \textbf{Access} \\
\midrule
GPT Luna + Websearch        & \texttt{gpt-5.6-luna}                  & OpenAI Responses API \\
GPT Sol + Websearch        & \texttt{gpt-5.6-sol}                  & OpenAI Responses API \\
Claude Sonnet 5 + Websearch & \texttt{claude-sonnet-5}               & Anthropic Messages API \\
Claude Opus 5 + Websearch & \texttt{claude-opus-5-5}               & Anthropic Messages API \\
Gemini Deep Research         & \texttt{deep-research-preview-04-2026} & Gemini Interactions API \\
DR Tulu                      & \texttt{rl-research/DR-Tulu-8B}        & local (vLLM + MCP server) \\
QUEST                        & \texttt{QUEST-35B-RL}                  & local \\
\bottomrule
\end{tabular}
\label{tab:system-ids}
\end{table*}

\subsection{Relationship between \textbf{SmoothScore} and generation features}
Table~\ref{tab:system-length-evidence} contains report length and retrieved evidence by dataset and system and Table~\ref{tab:correlates} shows if such features can predict \textbf{SmoothScore} with and across systems. We find that while systems show connection to different features across systems these relationships are weak and vary in direction, with only report length showing a small association.

\begin{table}[t]
\centering
\small
\setlength{\tabcolsep}{4pt}
\caption{
Report length and retrieved evidence by dataset and system, over the reports scored by the \textbf{SmoothScore} judge. ``Report words'' and ``Evidence words'' are medians over reports; ``Docs/report'' is the mean number of retrieved documents per report. ``Evidence words'' is the total number of words in the retrieved documents available to the system. 
}
\label{tab:system-length-evidence}
\begin{tabular}{llrrr}
\toprule
Dataset & System & Report words & Docs/report & Evidence words \\
\midrule
\multirow{7}{*}{ResearcherBench}
 & Luna 5.6 & 1,828 & 8.6  & 77,898 \\
 & Sol 5.6  & 1,710 & 9.9  & 92,880 \\
 & Sonnet 5 & 2,125 & 21.6 & 155,148 \\
 & Opus 5.5 & 1,328 & 18.8 & 138,441 \\
 & QUEST    & 1,332 & 9.8  & 55,481 \\
 & DR Tulu  & 1,948 & 17.4 & 114,027 \\
 & Gemini   & 4,860 & 39.2 & 254,916 \\
\midrule
\multirow{7}{*}{CS2-Expert}
 & Luna 5.6 & 1,999 & 10.4 & 34,821 \\
 & Sol 5.6  & 1,514 & 10.8 & 69,483 \\
 & Sonnet 5 & 2,177 & 18.5 & 175,065 \\
 & Opus 5.5 & 1,431 & 19.7 & 232,211 \\
 & QUEST    & 1,210 & 8.6  & 46,954 \\
 & DR Tulu  & 2,156 & 17.8 & 92,667 \\
 & Gemini   & 5,122 & 48.6 & 253,233 \\
\midrule
\multirow{7}{*}{DRB-II}
 & Luna 5.6 & 3,886 & 14.6 & 82,187 \\
 & Sol 5.6  & 3,979 & 20.2 & 162,523 \\
 & Sonnet 5 & 3,432 & 32.4 & 192,348 \\
 & Opus 5.5 & 2,786 & 31.2 & 205,332 \\
 & QUEST    & 2,397 & 11.2 & 40,119 \\
 & DR Tulu  & 2,884 & 14.4 & 110,388 \\
 & Gemini   & 4,844 & 32.7 & 235,720 \\
\midrule
\multirow{7}{*}{ResearchRubrics}
 & Luna 5.6 & 3,128 & 10.0 & 60,175 \\
 & Sol 5.6  & 2,736 & 11.9 & 83,795 \\
 & Sonnet 5 & 3,031 & 28.6 & 199,191 \\
 & Opus 5.5 & 2,010 & 22.7 & 134,858 \\
 & QUEST    & 1,930 & 9.6  & 52,607 \\
 & DR Tulu  & 2,732 & 15.0 & 106,390 \\
 & Gemini   & 4,687 & 31.2 & 196,485 \\
\bottomrule
\end{tabular}%

\end{table}

\begin{table}[h]
\centering\small
\renewcommand{\tabcolsep}{1.4mm}
\caption{Correlation of report and claim level features with \textsc{SmoothScore} across and within systems (Spearman $\rho$ and Pearson $r$). \emph{Across}: the mean over the 28 system--dataset. \emph{Within}: after subtracting each system--dataset mean. Systems that cite more documents or write longer claims smooth more, but within a system such relationships are weak and vary in direction. Only report length shows a small, consistent within-system association (Panel B)}
\label{tab:correlates}
\begin{tabular}{lcccc}
\toprule
\multicolumn{5}{l}{\textit{A. Pooled}} \\
\midrule
& \multicolumn{2}{c}{\textbf{Across systems}} & \textbf{Within system} & \textbf{Raw pooled} \\
\cmidrule(lr){2-3}
\textbf{Variable} & $\rho$ & $r$ & $\rho$ & $\rho$ \\
\midrule
Cited documents   & $+0.37$ ($.056$) & $+0.59$ ($.001$) & $+0.03$ ($.35$)  & $+0.38$ \\
Report length     & $+0.24$ ($.23$)  & $+0.39$ ($.04$)  & $+0.12$ ($.001$) & $+0.33$ \\
Number of claims  & $+0.08$ ($.70$)  & $+0.10$ ($.63$)  & $-0.02$ ($.61$)  & $0.00$ \\
Claim length      & $+0.43$ ($.02$)  & $+0.56$ ($.002$) & $+0.02$ ($.16$)  & $+0.12$ \\
\bottomrule
\end{tabular}

\vspace{0.8em}

\begin{tabular}{lccccccc}
\toprule
\multicolumn{8}{l}{\textit{B. Within-system correlations by system}} \\
\midrule
\textbf{Variable} & \textbf{Luna} & \textbf{Sol} &
\textbf{Sonnet~5} & \textbf{Opus~5.5} &
\textbf{QUEST} & \textbf{DR Tulu} & \textbf{Gemini} \\
\midrule
Cited documents
& +0.14 (.12) & +0.27 (.003) & +0.03 (.77) & 0.00 (.96)
& -0.14 (.11) & -0.19 (.03) & +0.09 (.35) \\

Report length
& +0.17 (.06) & +0.14 (.12) & +0.12 (.17) & +0.14 (.12)
& +0.02 (.81) & +0.16 (.08) & +0.06 (.50) \\

Number of claims
& -0.04 (.69) & +0.05 (.57) & 0.00 (.96) & +0.16 (.08)
& -0.11 (.22) & +0.01 (.90) & -0.23 (.01) \\

Claim length
& -0.04 (.30) & -0.01 (.85) & +0.08 (.02) & +0.06 (.09)
& -0.09 (.009) & +0.08 (.02) & +0.02 (.54) \\
\bottomrule
\end{tabular}
\end{table}

\section{\textsc{SmoothScore} Prompts}

\subsection{Claim Generation}
\label{sec:claim-generation}
\begin{prompt}[title={\footnotesize\texttt{Prompt \thetcbcounter: Main Conclusion Extraction}}, label=prompt:main_conclusion_extraction]

\promptsubsection{System Prompt}

You identify the MAIN CONCLUSIONS a research report reaches in answering
a question -- the handful of load-bearing verdicts a reader walks away
with -- and return each one as a COMPLETE verbatim passage from the
report. You are given the question and the report's answer.

BE SELECTIVE. Extract only the report's main conclusions: the thesis-level takeaways that directly answer the question. A typical report has roughly 4-8 of these. Favor the verdicts the whole report builds toward over the individual findings that support them.

UNIT OF EXTRACTION -- the complete conclusion, NOT an atom:

1. Lift each conclusion WHOLE: typically 1-3 consecutive sentences (roughly 15-80 words). The claim MUST include, whenever they appear in the same or immediately adjacent sentences: every hedge and qualifier; the flip-side, contrast, or counterpart condition ("; it does not ...", "However, ...", "but ..."); and the stated scope or dimension the verdict applies to. Emitting the assertive half of a hedged or paired statement is the worst possible error -- if the pieces cannot be included coherently, skip the conclusion entirely.

2. CARRY THE SCOPE. When a verdict's scope lives just OUTSIDE its sentence -- in the immediately governing heading, a dimension label, or the sentence before -- stitch that report wording in front (join stitched spans naturally; the heading text is report wording). NEVER present a per-dimension or per-condition winner as a global verdict.

3. NEVER extract a heading, list label, or table fragment as a claim, even when it is bolded and reads like a verdict. Headings point to claims; take the claim from the body text.

VOICE -- verbatim about the assertion, silent about the sourcing:

4. REUSE THE REPORT'S OWN WORDING at the report's own strength. Keep every hedge, qualifier, comparative, superlative, and number exactly as written. Never strengthen, weaken, or paraphrase the assertion.

5. STRIP ATTRIBUTION AND CITATION FRAMING: drop "X's experiments found that", "according to ...", "studies show", citation markup, and markdown link syntax. KEEP a system/method name only when the claim is ABOUT that system.

6. STRIP DISCOURSE CONNECTIVES at the very start of a claim ("On the other hand,", "In summary,", "However,", "Overall,") -- but connectives INSIDE the claim stay; they are doing real work joining the verdict to its counterpart.

WHAT COUNTS AS A CONCLUSION -- all three required:

- A conclusion the report draws, not a building block: a result, comparison, judgment, trade-off, trend, or recommendation that answers the question.

- Substantive, not a bare definitional or mechanistic statement.

- Contentful and checkable: it asserts something specific enough to check against evidence. A forecast or recommendation counts, but the claim must preserve the report's framing of it as a forecast/recommendation.

GRANULARITY -- one claim per distinct conclusion or thread. Different claims must not restate the same verdict; together they should cover each side's main strengths, the main limitations or trade-offs, and the overall verdict if the report gives one.

SKIP: background, definitions, setup, motivation, restatements of the question, section framing, individual supporting findings that roll up into a larger verdict.

OUTPUT -- return only JSON with this exact schema:

\texttt{\{"claims": [\{"text": "\{\{\paramnorm{Claim Text}\}\}",
"source\_text": "\{\{\paramnorm{Verbatim Source Span(s)}\}\}"\}]\}}

For each claim:

- "text": the complete conclusion passage, following the rules above. Plain prose -- no markdown, no citation markup, no list numbering.

- "source\_text": the exact verbatim span(s) of the answer the claim was lifted from, copied character-for-character (citation markup and markdown syntax may be dropped). If the claim stitches spans (e.g. heading scope + body), join them with " ... " in report order. Each span must be findable in the answer by string search.

FINAL CHECKLIST -- verify EVERY claim's "text" passes all of these; rewrite it if not:

- [ ] Complete sentence(s) -- never a fragment, bullet label, heading, or table cell.

- [ ] Includes the local hedges, scope, and flip-side that appear in or immediately beside the source passage.

- [ ] No attribution framing, no markdown
(\texttt{**}, \texttt{*}, \texttt{-}, \texttt{\#},
\texttt{>}, \texttt{|}), no citation markup, no URLs, no list numbering.

- [ ] The assertion is the report's wording at the report's strength -- qualifiers, comparatives, numbers unchanged.

If the answer states no checkworthy conclusions, return \{"claims": []\}.

\medskip
\textbf{User Prompt:}

Example ID: \{\{\paramnorm{Example ID}\}\}

Question: \{\{\paramnorm{Question}\}\}

Extract the MAIN CONCLUSIONS of the following final\_response -- the
handful of load-bearing, evidence-based verdicts that answer the
question (typically 3--7). For each, return the claim text (the
assertion in the report's own words at the report's own strength, with
all attribution, citation, discourse-connective, and markdown framing
STRIPPED, per the voice rules and final checklist) and the verbatim
source\_text span(s) it was lifted from. Use the question to decide
which conclusions are load-bearing.

final\_response:

\{\{\paramnorm{Full Report Text}\}\}

\end{prompt}

\subsection{Supporting Pass}
\label{sec:supporting_pass}
\begin{prompt}[title={\footnotesize\texttt{Prompt \thetcbcounter: Supporting Evidence Extraction}},
label=prompt:supporting_evidence]
\textbf{System Prompt:}

You extract from a single source DOCUMENT every distinct piece of information that SUPPORTS
a single CLAIM from a research report, or a specific part of it -- makes what the claim
actually says more plausible or provides a direct instance/evidence for it.

Extract ONLY supporting evidence. Ignore evidence against the claim entirely; another
process handles that.

\begin{itemize}
    \item POLARITY: support means supporting what the claim ACTUALLY ASSERTS. If the claim
    asserts a limitation or a negative (e.g.\ ``X is unreliable'', ``X scales worse'',
    ``X does not generalize''), then evidence of X failing/underperforming is supporting,
    and a result showing X working well is NOT supporting -- skip it.

    \item STATED CONCESSIONS ARE PART OF THE CLAIM: evidence confirming a limitation,
    condition, or concession the claim itself states supports the claim as stated --
    include it, with the concession named in the aspect field.
\end{itemize}

An item is one distinct FINDING: a specific result, experiment, measurement, or stated
conclusion. Each item is an object:

\begin{verbatim}
  {"quote": "<the shortest verbatim excerpt that carries this finding>",
   "aspect": "<the specific assertion WITHIN the claim this finding bears
               on, quoted or near-verbatim from the claim>",
   "note": "<one sentence: how this finding bears on that assertion>",
   "strength": <1-5>}
strength = how directly and specifically this item bears on what the
claim asserts:
  1 = tangential  2 = weak/indirect  3 = moderate: relevant but partial
  or hedged
  4 = strong: direct, on-point  5 = decisive: exactly what the claim
  asserts (or a clear counterexample to it).
\end{verbatim}

\textbf{Critical rules:}
\begin{itemize}
    \item Judge ONLY against this one document. No outside knowledge, no other documents.

    \item Read the WHOLE document; relevant evidence can sit in any section.

    \item ONE item per finding. If the document states the same finding several ways
    (abstract, body, conclusion), pick the single best quote for it -- do not list
    restatements or different facets of the same result as separate items.

    \item Do NOT include items that only establish setting or applicability without an
    outcome (e.g.\ ``the experiments use a 1.5B model''). That is context, not evidence; if
    it matters, mention it inside the note of a real item.

    \item An empty list is a perfectly good answer -- do NOT invent items. Merely being
    about the same topic is NOT evidence; every item must actually move belief about the
    claim in the direction you were asked for.

    \item MECHANISM: only count evidence that matches what the claim specifically says
    (e.g.\ if the claim says ``distillation from a LARGER model'', a result using a
    same-size or self-distilled model does not bear on that specific mechanism). The aspect
    field must name the exact assertion addressed -- if you cannot point to one, the item
    does not belong.

    \item Quotes must be copied verbatim from the document.
\end{itemize}

Return only JSON: \texttt{\{"items": [ \ldots ]\}}

\medskip
\textbf{User Prompt:}

Question (context only, not evidence): \{\{\paramnorm{Question}\}\}

Claim: \{\{\paramnorm{Claim}\}\}

Document:\\
Title: \{\{\paramnorm{Document Title}\}\}\\
URL: \{\{\paramnorm{Document URL}\}\}

Content:\\
\{\{\paramnorm{Full Document Text}\}\}

Extract every distinct piece of supporting evidence and return only the JSON object.
\end{prompt}

\subsection{Skeptic Pass}
\label{sec:skeptic_pass}
\begin{prompt}[title={\footnotesize\texttt{Prompt \thetcbcounter: Tagged Skeptical Evidence Extraction}},
label=prompt:skeptical_evidence]
\textbf{System Prompt:}

You extract from a single source DOCUMENT every distinct piece of information that a
SKEPTICAL reader would weigh against a single CLAIM from a research report, and you TAG
each item by its relation to the claim AS STATED. Extract only this skeptical side; a
separate process handles supporting evidence.

Read the claim carefully first: its actual assertions, its scope, its comparatives, and
every limitation, condition, or concession it ALREADY STATES or hedges over.

\medskip
TAGS:

\begin{itemize}
    \item \texttt{"counter"}: the finding cuts against what the claim ACTUALLY ASSERTS --
    contradicts it, provides a counterexample, limits its scope beyond what it admits,
    shows the asserted effect weaker or more conditional than asserted, or reports a null
    or mixed result where the claim implies a clear one. POLARITY: if the claim asserts a
    limitation or a negative (``X is unreliable''), a result showing X working WELL is a
    counter, and evidence of X failing is NOT. A soft hedge does not concede everything:
    ``X is usually best'' still asserts usually -- evidence X OFTEN loses is a counter;
    evidence X merely SOMETIMES loses is not.

    \item \texttt{"concession-match"}: the finding confirms, instantiates, or quantifies a
    limitation, condition, or concession the claim ITSELF states or hedges over
    (``capacity-sensitive'', ``results vary'', ``except in short contexts''). These do NOT
    count against the claim -- they are extracted so a later audit can check whether the
    report states the substance behind its own hedges.

    \item \texttt{"rival-win"}: a success of ANOTHER approach reported in that approach's
    own terms, where the claim does NOT rank over or exclude that approach. If the claim
    DOES rank or exclude (``strongest'', ``most practical'', ``X remains ahead of Y''), a
    rival's success is a \texttt{"counter"}, not a \texttt{"rival-win"}.
\end{itemize}

A carefully scoped claim often has NO counter items in a given document -- an empty counter
set is a common and correct outcome; do not stretch. But still extract concession-match and
rival-win items where they exist.

\begin{itemize}
    \item STRENGTH for \texttt{"counter"} = how much the finding should CONSTRAIN the claim
    as stated: 5 = directly contradicts the central assertion or shows it failing within its
    own stated scope; 4 = a hard limit or condition on the main assertion that the claim
    does not state; 3 = a meaningful qualification; 2 = weak/indirect pressure;
    1 = tangential.

    \item STRENGTH for \texttt{"concession-match"} = how load-bearing the conceded condition
    is: 5 = the concession hides a result that would change most readers' decisions;
    3 = a substantive but secondary condition; 1 = a trivial caveat.

    \item STRENGTH for \texttt{"rival-win"} = how strong the rival result is in its own
    terms (controlled comparison $>$ own-harness leaderboard $>$ assertion).
\end{itemize}

An item is one distinct FINDING: a specific result, experiment, measurement, or stated
conclusion. Each item is an object:

\begin{verbatim}
  {"quote": "<the shortest verbatim excerpt that carries this finding>",
   "aspect": "<the specific assertion WITHIN the claim this finding bears
               on, quoted or near-verbatim from the claim>",
   "note": "<one sentence: how this finding bears on that assertion>",
   "tag": "counter|concession-match|rival-win",
   "strength": <1-5>}
strength = how directly and specifically this item bears on what the
claim asserts:
  1 = tangential  2 = weak/indirect  3 = moderate: relevant but partial
  or hedged
  4 = strong: direct, on-point  5 = decisive: exactly what the claim
  asserts (or a clear counterexample to it).
\end{verbatim}

\textbf{Critical rules:}
\begin{itemize}
    \item Judge ONLY against this one document. No outside knowledge, no other documents.

    \item Read the WHOLE document; relevant evidence can sit in any section.

    \item ONE item per finding. If the document states the same finding several ways
    (abstract, body, conclusion), pick the single best quote for it -- do not list
    restatements or different facets of the same result as separate items.

    \item Do NOT include items that only establish setting or applicability without an
    outcome (e.g.\ ``the experiments use a 1.5B model''). That is context, not evidence; if
    it matters, mention it inside the note of a real item.

    \item An empty list is a perfectly good answer -- do NOT invent items. Merely being
    about the same topic is NOT evidence; every item must actually move belief about the
    claim in the direction you were asked for.

    \item MECHANISM: only count evidence that matches what the claim specifically says
    (e.g.\ if the claim says ``distillation from a LARGER model'', a result using a
    same-size or self-distilled model does not bear on that specific mechanism). The aspect
    field must name the exact assertion addressed -- if you cannot point to one, the item
    does not belong.

    \item Quotes must be copied verbatim from the document.
\end{itemize}

Return only JSON: \texttt{\{"items": [ \ldots ]\}}

\medskip
\textbf{User Prompt:}

Question (context only, not evidence): \{\{\paramnorm{Question}\}\}

Claim: \{\{\paramnorm{Claim}\}\}

Document:\\
Title: \{\{\paramnorm{Document Title}\}\}\\
URL: \{\{\paramnorm{Document URL}\}\}

Content:\\
\{\{\paramnorm{Full Document Text}\}\}

Extract every distinct piece of countervailing evidence and return only the JSON object.
\end{prompt}

\subsection{Aspect generation}
\label{sec:Aspect_generation}
\begin{prompt}[title={\footnotesize\texttt{Prompt \thetcbcounter: Document-Level Aspect Induction}},
label=prompt:aspect_induction]
\textbf{System Prompt:}

You are an analyst that identifies the different aspects that a single
scientific research paper ACTUALLY DISCUSSES which bear on evaluating a
claim. For each aspect, provide a brief all-lowercase label (3--8 words),
a one-sentence description of what THIS DOCUMENT says about it and why
it is significant for addressing the claim, and 6 keywords typically
used to describe that aspect.

Rules (corpus-grounded, claim-relative):
\begin{itemize}
    \item Only include aspects this document actually contains
    substantive content on---do not invent aspects from general
    knowledge of the topic.
    \item Frame each aspect as a dimension bearing on THIS claim
    (e.g., ``task-type dependence of the comparison''), not a generic
    subtopic of the field.
    \item Return as many aspects as the document genuinely supports:
    typically 2--3.
\end{itemize}

One aspect is a good answer for a narrowly relevant document, and an
EMPTY LIST is the correct answer for a document with nothing substantive
on this claim. Do not pad to a count.

Return only JSON:
\texttt{\{"aspect\_list": [\{"aspect\_label": "...",
"aspect\_description": "...", "aspect\_keywords": ["..."]\}]\}}.\\

\medskip
\textbf{User Prompt:}

CLAIM: \{\{\paramnorm{Claim}\}\}

DOCUMENT \#\{\{\paramnorm{Document Index}\}\}
(title: \{\{\paramnorm{Document Title}\}\}):\\
\{\{\paramnorm{Full Document Content}\}\}

Return only the JSON object.
\end{prompt}

\subsection{Aspect Merge}
\label{sec:Aspect_merge}
\begin{prompt}[title={\footnotesize\texttt{Prompt \thetcbcounter: Aspect Hierarchy Consolidation}},label=prompt:aspect_consolidation]
\textbf{System Prompt:}

You are an analyst consolidating per-document aspect proposals into ONE
aspect hierarchy for evaluating a claim. You will identify 4--5
top-level aspects that the documents' proposals are collectively
considering when determining the answer to the claim. Add subaspects
ONLY where the corpus genuinely subdivides an aspect into distinct,
separately-evidenced points---a top-level aspect with NO subaspects is a
good and common answer. Aim for roughly 4--8 nodes in the whole
hierarchy, not the maximum. Each subaspect must name what distinguishes
it from its siblings; if you cannot, merge them.

Rules:
\begin{itemize}
    \item Corpus-grounded: every aspect/subaspect must be traceable to
    the per-document proposals; carry over the source document indexes
    into \texttt{"source\_docs"}. Do not invent nodes no document
    proposed.
    \item Claim-relative: keep labels framed as dimensions bearing on
    the claim.
    \item Merge duplicates across documents; prefer the most specific
    phrasing proposed.
    \item A subaspect is a more specific point underneath its parent
    that would be considered when specifically addressing the claim.
\end{itemize}

Return only JSON:
\texttt{\{"aspects": [\{"label": "...", "description": "...",
"source\_docs": [1,2], "subaspects": [\{"label": "...",
"description": "...", "source\_docs": [3]\}]\}]\}}.\\

\medskip
\textbf{User Prompt:}

CLAIM: \{\{\paramnorm{Claim}\}\}

PER-DOCUMENT ASPECT PROPOSALS:

[doc 1] \{\{\paramnorm{Aspect Label}\}\}:
\{\{\paramnorm{Aspect Description}\}\}

[doc 1] \{\{\paramnorm{Aspect Label}\}\}:
\{\{\paramnorm{Aspect Description}\}\}

[doc 3] \{\{\paramnorm{Aspect Label}\}\}:
\{\{\paramnorm{Aspect Description}\}\}

\ldots

Return only the JSON object.
\end{prompt}

\subsection{Evidence Assignment}
\label{sec:Evidence_Assignment}
\begin{prompt}[title={\footnotesize\texttt{Prompt \thetcbcounter: Evidence-to-Aspect Assignment}},label=prompt:evidence_assignment]
\textbf{System Prompt:}

You will be provided with evidence quotes extracted from source documents.
Each quote was extracted as evidence about a claim. Please select the
suitable aspect each quote talks about from the provided aspect list.

LEAF-ONLY RULE: a node that has sub-aspects is a container only. If the
quote belongs in such an aspect's territoryts sub-aspects (thebest-fitting
one), never the parent. Only aspects with no sub-aspects can receive quotes
directly.

Special destinations:
\begin{itemize}
    \item ``ROOT'' if the quote bears oher than any one aspect(overall
    verdicts, provenance of the whole comparison, conditions spanning
    aspects)ROOT is a deliberate destinat
    \item ``UNASSIGNED'' only if the quote fits no listed aspect and is not
    whole-claim evidence.
\end{itemize}

Assignment is topical: judge what the quote is ABOUT, never whether it
supports or opposes the claim.

Each quote is prefixed with its source document (index and title). Use it
as cona fragment (a table row, a bare number, a one-line result) is about
whatever that document discusses at that point -- file it by its
document's topic, not by the words in the fragment alone.

Return only JSON:
\texttt{\{"assignmen "node": "<node id like A1 orA2.3 or ROOT or
UNASSIGNED>"\}]\}}.

\medskip
\textbf{User Prompt:}

CLAIM: \{\{\paramnorm{Claim}\}\}

ASPECT LIST:

A1: \{\{\paramnorm{Aspect Label}\}\}

A1.1: \{\{\paramnorm{Aspect Label \(>\) Subaspect Label}\}\}

A2: \{\{\paramnorm{Aspect Label}\}\}

\ldots

EVIDENCE:

[0] (doc 3 --- \{\{\paramnorm{Title, 90 Charsned, 400 Chars}\}\})

[1] (doc 7 --- \{\{\paramnorm{Title}\}\})
\{\{\paramnorm{Quote}\}\}
\end{prompt}

\subsection{\textsc{SmoothScore} Judge}
\label{sec:Smoothscore_judge}

\begin{prompt}[title={\footnotesize\texttt{Prompt \thetcbcounter: Evidence-Smoothing Judgment}},label=prompt:smoothing_judgment]
\medskip
\textbf{System Prompt:}
You are an evidence-smoothing judge. You assess how faithfully a claim compresses the evidence supplied to it — not whether the claim is well-written, complete, or true by outside standards.

\medskip
\textbf{User Prompt:}
\# Inputs

Research question:
\{\{QUESTION\}\}

Claim under evaluation:
\{\{CLAIM\}\}

Evidence by aspect:
\{\{ASPECTS\_AND\_EVIDENCE\}\}

Each evidence item may include: a source identifier, an exact quotation, a generated context note, and a retrieval-pass label ("support-pass candidate" / "countervailing-pass candidate").\\

\# What you're measuring

Evidence smoothing: the claim collapses material heterogeneity, uncertainty, conditionality, or disagreement in the evidence, producing a conclusion broader, more certain, more universal, or more decisive than the evidence proportionately warrants.

Not in scope: coverage (the claim need not mention every finding), writing quality, or accuracy against knowledge outside the supplied evidence.\\

\# Ground rules for reading the evidence

1. **The supplied evidence is the entire universe.** No outside knowledge.\\
2. **Retrieval labels describe provenance, not verified stance.** Interpret each item's actual content independently — a support-pass item can fail to support the claim; a countervailing-pass item can support, qualify, or be irrelevant to it.\\
3. **Item counts and pass labels are not evidence of consensus or its absence.** Never infer balance, or lack of it, from how many items landed in each pass.\\
4. **Context notes clarify; they don't add evidence.** Ground judgments in the quotation. Preserve the claim’s actual scope, certainty, and role in answering the question. Do not weaken a strong assertion into a vague statement of usefulness, or strengthen a qualified claim into an unconditional guarantee. \\
5. **Weigh by relevance, directness, applicability, and any quality signal stated in the text — not by quote count.** Restatements of the same finding are one data point, not several.\\
6. **The claim's own logical force sets what can defeat it:**
   - "X can work" — isolated failures of X don't count against it.
   - "X always / necessarily works" — one sound counterexample is decisive.
   - "X generally works / dominates" — isolated counterexamples aren't enough; the evidence must speak to prevalence or the overall comparison.
   - "X is promising" — a rival promising method doesn't refute it.
   - "X is comparatively best" — comparable alternatives are directly relevant.
   - A claim need not preempt exceptions its own wording already allows.\\
7. **Absence of support \(\neq\) contradiction**, but an unsupported specific assertion is still unwarranted — weight this by whether it's peripheral or central to the takeaway.\\
8. **For multi-part claims, weight problems by their effect on the overall takeaway.** Don't average proposition-level defects or default to the score triggered by the single worst one.\\

\# Procedure

Work through these in order. Your response's structure follows this order (see Output).

1. **Claim's contribution.** What is the claim doing relative to the question — explaining a possibility, asserting a general pattern, reporting a result, making a recommendation? This determines what kind of evidence can defeat it.\\
2. **Substantive takeaway.** State the scope, certainty, comparison, and causal content the claim actually asserts, independent of phrasing. Don't pre-weaken an overstated claim before judging it.\\
3. **Positive evidential case.** What do the relevant findings, taken together, justify? Include reasonable inferences across findings. Flag anything the claim asserts that the evidence doesn't actually reach (a specific mechanism, magnitude, or comparison it never establishes), even if the surrounding evidence is generally favorable.\\
4. **Objections.** For each apparent counterpoint: does it address the same assertion, in the applicable setting? Distinguish a consequential restriction from a compatible failure, a different mechanism, or an unrelated limitation. Treat a finding as an outlier only when the evidence itself gives a basis for that, not because it's outnumbered or inconvenient.\\
5. **Net resolution.** Combine 3 and 4. Does the claim survive as stated, survive with a limitation attached, or require a substantially different conclusion?\\
6. Score via the minimum-repair test below. \\

\# Minimum-repair test

Determine the score by identifying the minimum semantic revision needed to make the claim faithful to the supplied evidence. Work through these levels in order and stop at the first one that fits:

- None: the claim can remain as written. Omitted material would only elaborate or add background. A further caveat could be written, but none is needed for fidelity.\\
- Append: a caveat or exception should be added for the claim to represent the evidence faithfully, but the original claim otherwise remains — its central scope, certainty, comparison, and conclusion stay substantially unchanged.\\
- Revise: the claim's central direction survives, but a material predicate, quantifier, scope, comparison, causal attribution, or confidence level must be changed. A reader who accepted the original would draw a materially different conclusion from the repaired version.\\
- Recast: the overall conclusion must be substantially narrowed, rebalanced, or made explicitly mixed. A recognizable part of the original takeaway survives, but not the conclusion as presented.\\
- Replace: the central takeaway cannot be preserved. A faithful account requires a materially different conclusion; adding qualifications or rebalancing is insufficient.\\

Do not select Append merely because an additional caveat could be written — select it only when a caveat is genuinely needed for faithful representation. Do not select Revise for a hedge the claim's own wording already implies (e.g. adding "generally" to a claim that elsewhere says "typically") — that is an Append.\\

\# Scoring rubric

Each repair level corresponds to exactly one score: \\

None \(\rightarrow\) 1 (Reasonable compression). The claim preserves important conditions and fairly represents relevant counterevidence. Omitted details do not meaningfully change the takeaway.\\

Append \(\rightarrow\) 2 (Minor nuance lost). Some conditions, limitations, or exceptions are omitted, but including them would leave the takeaway largely unchanged.\\

Revise \(\rightarrow\) 3 (Important qualification missing). An important condition, limitation, or exception is omitted, making the claim broader or more certain than warranted. The central takeaway remains defensible with that qualification.\\

Recast \(\rightarrow\) 4 (Major distortion). Substantial counterevidence or a decisive limitation is omitted or understated. The claim presents a clear conclusion where the evidence requires substantial narrowing, rebalancing, or a mixed conclusion.\\

Replace \(\rightarrow\) 5 (Fundamentally misleading). The claim fundamentally misrepresents what the evidence warrants — for example, by omitting both crucial conditions and strong counterevidence. The central takeaway needs replacing; adding qualifications alone cannot make it faithful. Direct contradiction is not required.\\

\# Deciding the score\\

Decide from the state of the evidence around the central takeaway, then confirm against the rubric above.\\

Only material bearing on the central takeaway counts here. Material that qualifies a subsidiary assertion — an example, a mechanism, a side-condition, an intensifier — does not move the score unless correcting it changes the takeaway itself.

The claim's hedges are part of its wording. “Likely”, “often”, “when X”, “reducing” set what has to be shown. Evidence that the claim is not uniformly or always true does not bear against a claim that never asserted universality. \\

- Nothing in the evidence bears against the takeaway, or what does is already allowed by the claim's own wording \(\rightarrow\) 1\\
- What bears against it is narrower in scope, less direct, or less consistent than what supports it — the takeaway holds with surety \(\rightarrow\) 2\\
- What bears against it is as direct and as broad as what supports it — the takeaway is arguably correct but contested \(\rightarrow\) 3\\
- What bears against it outweighs the support, or the two together leave the central question genuinely two-sided — a faithful account is mixed \(\rightarrow\) 4\\
- The evidence never establishes the takeaway, or establishes something materially different \(\rightarrow\) 5\\

The repair labels — none, append, revise, recast, replace — describe the consequence of the score and are reported for auditing. They are not the test. Before assigning a score, identify a material qualification required by the evidence that is not already conveyed by the claim’s wording or context.\\

\# Output

Return a single JSON object with exactly these fields, in this order. No text before or after the JSON.

\begin{verbatim}
{
  "claim_role": "<one sentence: what kind of claim this is
  (possibility / general pattern / result / recommendation),
  and what that implies for what can defeat it>",

  "takeaway": "<the claim's actual scope, certainty, comparison,
  and causal content, stated plainly>",

  "positive_case": "<what the relevant evidence, taken together,
  establishes, and, if applicable, what the claim asserts beyond
  what the evidence actually reaches>",

  "objections": "<what the countervailing or qualifying evidence
  actually changes, if anything; use 'none material' if nothing
  meaningfully bears on the claim as scoped>",

  "net_resolution": "<does the claim survive as-is, survive with
  a limitation, or need substantial revision, and why, in 1-2
  sentences>",

  "repair": "none | append | revise | recast | replace",

  "score": 1
}
\end{verbatim}

Constraints:
- "repair" and "score" must correspond exactly: none=1, append=2, revise=3, recast=4, replace=5.
- "score" must be an integer, not a string.
- Every field is required. Do not add extra fields.
- Output must be valid JSON — no markdown code fences, no trailing commas, no comments.
\end{prompt}

\section{Judge Validation}
\label{app:judge-agreement}

Two authors independently scored 73 claims on the same 1--5 scale
(Table~\ref{tab:judge-agreement}). GPT-5, our judge of record, agrees with Annotator~1 as closely as Annotator~2 does (ordinal $\alpha$ 0.62 vs.\ 0.64). Gemini~2.5~Pro and Claude Sonnet~4.5 agree with the annotators less well than GPT-5 does ($\alpha$ 0.36 and 0.04), although they still rank claims similarly ($\rho$ 0.60 and 0.51).

\begin{table}[h]
\centering\small
\caption{Agreement with Annotator~1 on 73 claims (Krippendorff's $\alpha$ is ordinal). \emph{Exact}: identical score. \emph{Within 1}: scores differ by at most one point. GPT-5 agrees with Annotator~1 as closely as the second annotator does; other judges rank claims similarly but differ in calibration.}
\label{tab:judge-agreement}
\begin{tabular}{lcccc}
\toprule
\textbf{Judge} & \textbf{Exact} & \textbf{Within 1} & \textbf{Spearman $\rho$} &
\textbf{Krippendorff $\alpha$} \\
\midrule
Annotator 2              & 58\% & 96\% & 0.72 & 0.71 \\
GPT-5 (judge of record)  & 56\% & 93\% & 0.69 & 0.68 \\
Gemini 2.5 Pro           & 27\% & 74\% & 0.63 & 0.41 \\
Claude Sonnet 4.5        & 19\% & 56\% & 0.51 & 0.11 \\
\bottomrule
\end{tabular}
\end{table}

All judges preserve the same system ranking as the annotator (Luna $<$ DR~Tulu $<$ Gemini;
Table~\ref{tab:judge-system-means}), with Gemini~2.5~Pro and Sonnet~4.5 scoring
systematically higher.

\begin{table}[h]
\centering\small
\caption{Mean score per system on the 73 annotated claims, by judge. All judges rank the
systems in the same order as the annotator.}
\label{tab:judge-system-means}
\begin{tabular}{lcccc}
\toprule
\textbf{System} & \textbf{Annotator~1} & \textbf{GPT-5} & \textbf{Gemini 2.5 Pro} &
\textbf{Sonnet 4.5} \\
\midrule
Luna                 & 1.96 & 1.71 & 2.33 & 2.88 \\
DR Tulu              & 2.16 & 2.60 & 2.96 & 3.32 \\
Gemini Deep Research & 2.71 & 3.04 & 3.83 & 4.42 \\
\bottomrule
\end{tabular}
\end{table}

\section{Step by Step outputs of the \textsc{SmoothScore} pipeline}

Table~\ref{tab:running-example} shows a running example of our framework processing a claim. Each step and the corresponding output for the example is shown. 

\begin{longtable}{p{0.19\textwidth}p{0.75\textwidth}}
\caption{Step-by-step application of the evidence-smoothing pipeline to
a claim about reinforcement learning with verifiable rewards.}
\label{tab:running-example}\\
\toprule
\textbf{Pipeline step} & \textbf{Output for the example} \\
\midrule
\endfirsthead

\multicolumn{2}{l}{\small\textit{Table~\ref{tab:running-example}
continued}}\\
\toprule
\textbf{Pipeline step} & \textbf{Output for the example} \\
\midrule
\endhead

\bottomrule
\endfoot

\textbf{Input question}
&
Will RL-based methods become the mainstream approach for improving
reasoning models after DeepSeek-R1, and can reward models and tree
search for chain-of-thought post-training be abandoned?
\\
\midrule

\textbf{1. Claim extraction}
&
\emph{``RL---especially RL with verifiable rewards---will become a
mainstream component of reasoning-model training.''}
\\
\midrule

\textbf{Document reconstruction}
&
The system reconstructs the full text of the documents cited by the
report, including work on DeepSeek-R1, verifier gaming, test-time
scaling, AlphaProof, AlphaGeometry2, and surveys of RL for reasoning.
\\
\midrule

\textbf{2a. Supporting-evidence extraction}
&
The supporting pass extracts findings that substantiate the claim as
stated. For example:

\medskip
\emph{``Figure 1(a) depicts the performance trajectory of
DeepSeek-R1-Zero on the AIME 2024 benchmark throughout the RL training
process, where the average pass@1 score on AIME 2024 shows a significant
increase, jumping from an initial 15.6\% to 77.9\%.''}

\medskip
\emph{``As reinforcement Learning with Verifiable Rewards (RLVR) has
become the dominant paradigm for scaling reasoning capabilities in
LLMs.''}

\medskip
Across all claim--document pairs, the pass extracts \(61\) supporting
items.
\\
\midrule

\textbf{2b. Countervailing-evidence extraction}
&
The countervailing pass extracts findings that contradict, weaken, or
materially restrict the claim. For example:

\medskip
\emph{``We show that training on only 1,000 samples with next-token
prediction and controlling thinking duration via a simple test-time
technique we refer to as budget forcing leads to a strong reasoning
model that scales in performance with more test-time compute.''}

\medskip
\emph{``The rule-based RL training stage of DeepSeek-R1-Zero is narrowly
focused on reasoning tasks, resulting in limited performance in broader
areas such as writing and open-domain question answering.''}

\medskip
Across all claim--document pairs, the pass extracts \(77\)
countervailing items.
\\
\midrule

\textbf{3a. Local aspect induction}
&
For each claim--document pair, the system proposes dimensions discussed
by that document that bear on evaluating the claim. For example,
DeepSeek-R1 proposes an aspect concerning \emph{performance gains on
objectively verifiable reasoning tasks}, while the test-time-scaling
paper proposes an aspect concerning \emph{non-RL alternatives for
improving reasoning}.
\\
\midrule

\textbf{3b. Aspect consolidation}
&
Overlapping document-level proposals are merged into the following
claim-level hierarchy:

\begin{itemize}
    \item \textbf{A1: Empirical effectiveness and domain scope of RL
    with verifiable rewards}
    \begin{itemize}
        \item \textbf{A1.1:} Large gains on tasks with objective checks
        \item \textbf{A1.2:} Boundaries and alternatives where RL is
        unnecessary or underperforms
    \end{itemize}

    \item \textbf{A2:} Dependence on verifier reliability and design for
    safe, scalable use

    \item \textbf{A3:} Practicality and adoption readiness of RLVR in
    training pipelines
    \begin{itemize}
        \item \textbf{A3.1:} Operationalization and community uptake
        \item \textbf{A3.2:} Resource costs and sample-efficiency
        trade-offs
    \end{itemize}
\end{itemize}
\\
\midrule

\textbf{4. Evidence assignment}
&
Supporting and countervailing items are assigned to the most specific
applicable aspect. For example:

\medskip
\textbf{A1.1: Large gains on tasks with objective checks}

\textit{Supporting evidence:}
\begin{itemize}
    \item DeepSeek-R1-Zero's AIME 2024 pass@1 score rises from
    \(15.6\%\) to \(77.9\%\) during RL training.
    \item AlphaProof learns to produce formal proofs through RL over
    millions of auto-formalized problems.
\end{itemize}

\textit{Countervailing evidence:}
\begin{itemize}
    \item AlphaGeometry2 achieves an \(84\%\) solve rate using a pipeline
    centered on language modeling, symbolic reasoning, and synthetic
    data rather than RL.
    \item AlphaProof's demonstrated successes remain concentrated in
    competition mathematics.
\end{itemize}

This node contains \(31\) supporting and \(9\) countervailing
items. Across the hierarchy, this is the only node in which supporting
items predominate; the remaining dimensions contain substantial
countervailing evidence.
\\
\midrule

\textbf{5. Smoothing judgment}
&
\textbf{Claim role:} General forecast about a field-wide training trend;
it can be defeated by evidence that RL with verifiable rewards is
unlikely to be widely adopted or that strong, persistent limitations
preclude mainstream use.

\medskip
\textbf{Takeaway:} Reinforcement learning---particularly when rewards
are verifiable---will be widely adopted as a standard part of training
pipelines for reasoning models, though not necessarily the only or
exclusive approach.

\medskip
\textbf{Objections:} Important constraints temper the breadth of the
forecast. RL scales poorly where no reliable verifier exists; it can
underperform or fail to improve smaller models; compute and evaluation
costs limit practicality; and verifier imperfections introduce
shortcutting and misalignment risks. Strong non-RL alternatives can
also match or surpass leading RL-trained systems. These findings show
that RLVR will not be universally applicable or singularly dominant
across reasoning tasks.

\medskip
\textbf{Net resolution:} The claim largely holds but needs an explicit
limitation: RL with verifiable rewards is poised to be mainstream for
verifier-evaluable settings, yet adoption will be constrained by
verifier availability, safety and shortcut risks, model and compute
regimes, and viable non-RL alternatives that will continue to coexist.

\medskip
\textbf{Repair:} \textsc{Append}
\hspace{1em}
\textbf{Smoothing score:} \(2\)
\\

\end{longtable}

\section{Report regeneration}

\subsection{Setup for WebSearch}

\label{sec:websearch_no_instructions}
\begin{prompt}[title={\footnotesize\texttt{Prompt \thetcbcounter: Report Generation with Web Search}},
label=prompt:websearch_no_instructions]
\textbf{System Prompt:}

You are a professional researcher producing a structured, citation-backed
report in response to the user's question. Ground every factual claim in
the web sources you retrieve and cite them inline. Be precise and
analytical. Prefer recent,
authoritative sources.\\

\medskip
\textbf{User Prompt:}

\{\{\paramnorm{Question}\}\}
\end{prompt}

\subsection{Setup for WebSearch, with Instructions}

\label{sec:websearch_with_instructions}
\begin{prompt}[title={\footnotesize\texttt{Prompt \thetcbcounter: Report Generation with Web Search (with Calibration Instructions)}},
label=prompt:websearch_with_instructions]
\textbf{System Prompt:}

You are a professional researcher producing a structured, citation-backed
report in response to the user's question. Ground every factual claim in
the web sources you retrieve and cite them inline. Be precise and
analytical. Prefer recent, authoritative sources.

\medskip
Readers will rely on your report to understand what the evidence does and
does not establish, so keep each conclusion calibrated to the sources:

\medskip
State the setting or conditions in which  not
generalize beyond them.\\
Where the sources disagree with each other
than choosing a side silently.\\
Do not combine separate findings into a se
sources establish.\\
Avoid conclusive language unless the sour

\medskip
If the sources do not support a conclusion, say so.\\

\medskip
\textbf{User Prompt:}

\{\{\paramnorm{Question}\}\}
\end{prompt}

\subsection{Setup for Docs in Context, with Instructions}

\label{sec:docs_with_instructions}
\begin{prompt}[title={\footnotesize\texttt{Prompt \thetcbcounter: Report Generation from Provided Documents (with Calibration Instructions)}}, label=prompt:docs_with_instructions]
\textbf{System Prompt:}

You are a professional researcher producing a structured, citation-backed
report. The complete source documents are provided below; they are the
ONLY evidence you may use. Ground every factual claim in them and cite by
document number, e.g. [doc 3].

\medskip
Readers will rely on your report to understand what the evidence does and
does not establish, so keep each conclusion calibrated to the documents:

\medskip
State the setting or conditions in which a result was shown, and do not
generalize beyond them.\\
Where the documents disagree with each other, say so explicitly rather
than choosing a side silently.\\
Do not combine separate findings into a stronger conclusion than the
documents establish.\\
Avoid conclusive language unless the documents clearly warrant it.

\medskip
If the documents do not support a conclusion, say so.\\

\medskip
\textbf{User Prompt:}

SOURCE DOCUMENTS

=== DOCUMENT 1 ===\\
URL: \{\{\paramnorm{Document URL}\}\}\\
TITLE: \{\{\paramnorm{Document Title}\}\}

\{\{\paramnorm{Document Text}\}\}

=== DOCUMENT 2 ===\\
URL: \{\{\paramnorm{Document URL}\}\}\\
TITLE: \{\{\paramnorm{Document Title}\}\}

\{\{\paramnorm{Document Text}\}\}

\ldots

=== QUESTION ===\\
\{\{\paramnorm{Question}\}\}

Write the report now, citing documents by
\end{prompt}

\subsection{ Setup for Docs in Context, No Instructions}
\label{sec:docs_no_instructions}
\begin{prompt}[title={\footnotesize\texttt{Prompt \thetcbcounter: Report Generation from Provided Documents (No Instructions)}}, label=prompt:docs_no_instructions]
\textbf{System Prompt:}

You are a professional researcher producing a structured, citation-backed
report. The complete source documents are provided below; they are the
ONLY evidence you may use. Ground every factual claim in them and cite by
document number, e.g. [doc 3].\\

\medskip
\textbf{User Prompt:}

SOURCE DOCUMENTS

=== DOCUMENT 1 ===\\
URL: \{\{\paramnorm{Document URL}\}\}\\
TITLE: \{\{\paramnorm{Document Title}\}\}

\{\{\paramnorm{Document Text}\}\}

=== DOCUMENT 2 ===\\
URL: \{\{\paramnorm{Document URL}\}\}\\
TITLE: \{\{\paramnorm{Document Title}\}\}

\{\{\paramnorm{Document Text}\}\}

\ldots

=== QUESTION ===\\
\{\{\paramnorm{Question}\}\}

Write the report now, citing documents by number.
\end{prompt}

\subsection{Improved claims under document engagement}
\label{sec:new_claims}
Table~\ref{tab:engagement-patterns} shows how engaging with documents can repair claims. We manually select cases where the same claims persist across setups but change in terms of scope and strength representing the cited documents accurately.  

\begin{table*}[p]
\centering
\scriptsize
\setlength{\tabcolsep}{3pt}
\renewcommand{\arraystretch}{1.1}
\caption{How smoothing changes when the same model is given the cited documents (Luna, ResearcherBench). Claims are extracted independently per report; each row shows the claim in each arm that reaches the same conclusion, with $\sigma$ in parentheses. ``--'' means no claim in that arm reaches the same conclusion.}
\label{tab:engagement-patterns}
\begin{tabular}{@{}
  >{\RaggedRight\arraybackslash}p{0.11\textwidth}
  >{\RaggedRight\arraybackslash}p{0.15\textwidth}
  >{\RaggedRight\arraybackslash}p{0.23\textwidth}
  >{\RaggedRight\arraybackslash}p{0.23\textwidth}
  >{\RaggedRight\arraybackslash}p{0.23\textwidth}
@{}}
\toprule
\textbf{Pattern} & \textbf{Question}
& \textbf{WebSearch ($\sigma$)} & \textbf{Docs in context ($\sigma$)} & \textbf{Docs + instructions ($\sigma$)} \\
\midrule

A result from one setup stated without its conditions
& Analyze the differences between the LIMO and S1 papers: objectives, methodology, key findings, contributions.
& LIMO appears stronger in broad benchmark performance, particularly on GPQA Diamond and MATH500. (4)
& LIMO reports higher scores than \emph{original} s1 on AIME24, MATH500 and GPQA, but the studies differ in evaluation details (greedy decoding vs.\ temperature-0) \ldots\ the comparison does not by itself establish a controlled head-to-head superiority claim. (1)
& LIMO is the more explanatory and curation-focused paper \ldots\ [doc 1]. (3) \\
\addlinespace

A result from one setup stated without its conditions
& How can we detect hallucinations in LLMs using their internal states, and what advantages does this offer over external methods?
& Lower latency and computational cost: internal detectors can operate in the same forward pass and use a lightweight probe. MIND targets real-time detection, and HIDE reports substantially lower computation time than multi-pass methods. (2)
& --
& Its principal advantages over external or output-only detection are earlier intervention, richer process-level information, reduced reliance on external knowledge bases, \emph{lower latency for single-pass methods}, and improved interpretability. (1) \\
\midrule

Separately supported fragments combined into one demonstrated conclusion
& Could transformer architectures be reimagined to process multimodal inputs (video/audio/text) as efficiently as text?
& The more promising principle is: use modality-specific perception, adaptive semantic tokenization, sparse cross-modal attention, persistent latent memory, and on-demand access to raw evidence. (4)
& The evidence suggests that multimodal Transformers \emph{may} need to be redesigned around asymmetric processing, learned compression, latent bottlenecks, adaptive tokenization, and selective cross-modal interaction. (3)
& The likely path \ldots\ is to redesign the architecture so that all modalities are converted into compact, queryable representations \ldots\ Whether that bottleneck can remain small enough \ldots\ \emph{remains unresolved by the supplied evidence}. (1) \\
\addlinespace

Separately supported fragments combined into one demonstrated conclusion
& How might federated learning combined with model distillation overcome technical and legal barriers in domains like healthcare?
& The most credible approach is therefore a layered architecture: local training + distillation + secure aggregation + differential privacy + strong governance and clinical validation. (3)
& The strongest approach is therefore layered: local data retention, carefully selected distillation, secure aggregation, differential privacy where appropriate \ldots\ and clinical validation. (3)
& The strongest approach is therefore layered: [same list] \ldots\ Such an architecture can substantially lower both technical and legal risk, \emph{but the supplied documents do not support claiming that it eliminates either category of barrier}. (1) \\
\midrule

Conclusive or superlative wording beyond what the sources state
& What is the evolution path of multimodal models from early visual representations to current multimodal LLMs, and what are the key breakthroughs?
& Language as a \emph{universal} interface for vision: natural language became both a source of supervision and a way to query visual representations. (3)
& CLIP's central breakthrough was language-grounded open-vocabulary recognition \ldots\ The approach also exposed important limitations: performance depended on prompt wording, and CLIP struggled with counting, fine-grained recognition, and out-of-distribution images [doc 11]. (1)
& CLIP and ALIGN demonstrated that web-scale paired text can train general visual representations, support retrieval, and produce zero-shot classifiers through natural language [docs 11, 13]. (1) \\
\addlinespace

Conclusive or superlative wording beyond what the sources state
& (same question)
& BLIP-2 made the connection more parameter-efficient. This was an \emph{important engineering breakthrough} because it showed that multimodal capability could be obtained without end-to-end retraining of every parameter. (3)
& --
& Flamingo and BLIP-2 showed that strong pretrained vision and language models can be connected through relatively small trainable modules rather than fully retrained end-to-end [docs 15--16]. (1) \\
\bottomrule
\end{tabular}
\end{table*}

\section{Document Engagement Details}
\label{app:document-engagement}

We measure how often each system opens the documents it cites. A cited link counts as \emph{opened} if the system retrieved that page, in any form, during generation. URLs are canonicalized before matching (arXiv \texttt{/abs/} and \texttt{/pdf/} variants merged;
tracking parameters and trailing slashes removed), and each cited link is counted once per dataset. Table~\ref{tab:engagement-all} reports all systems and datasets.

\begin{table}[h]
\centering\small
\renewcommand{\tabcolsep}{1.2mm}
\caption{Share of cited links opened, by system and dataset (opened / cited links). A link counts as opened if the system issued an open, fetch or visit for it. Gemini Deep Research exposes no browsing steps.}
\label{tab:engagement-all}
\begin{tabular}{lcccc}
\toprule
\textbf{System} & \textbf{ResearcherBench} & \textbf{CS2-Expert} & \textbf{DRB-II} & \textbf{ResearchRubrics} \\
\midrule
Luna 5.6             & 19.2\% (95/496)   & 13.3\% (34/256) & 9.3\% (27/291)  & 8.9\% (17/190) \\
Sol 5.6              & 27.7\% (169/611)  & 30.0\% (88/293) & 16.1\% (68/422) & 28.2\% (70/248) \\
QUEST                & 73.8\% (217/294)  & 64.9\% (63/97)  & 30.5\% (36/118) & 29.4\% (25/85) \\
Sonnet 5             & 2.3\% (34/1,458)  & 3.5\% (19/536)  & 1.1\% (8/750)   & 1.1\% (6/559) \\
Opus 5.5             & 0.9\% (12/1,277)  & 2.3\% (13/561)  & 1.3\% (10/751)  & 0.2\% (1/471) \\
DR Tulu              & 0.5\% (5/1,082)   & 0\% (0/445)     & 0\% (0/303)     & 0\% (0/290) \\
Gemini Deep Research & --                & --              & --              & -- \\
\bottomrule
\end{tabular}
\end{table}


\paragraph{Luna, Sol, Sonnet~5, Opus~5.5, and DR~Tulu.} Luna and Sol use the OpenAI Responses API web search tool, whose calls are recorded as search, page-open, or find-in-page actions \citep{openai2026websearch}. We count a cited link as opened if the returned search results contain that page's opened content, marked by a \texttt{Source: open(\ldots)} header. These markers capture every page-open and find-in-page action, as well as opens returned inside other calls' results, so counting the actions alone would miss about three quarters of them. Sonnet~5 and Opus~5.5 run with Anthropic's web search and web fetch tools. Search results return page content in encrypted form, so we cannot observe what the model sees from search alone, while web fetch retrieves a page's full text \citep{anthropic2026websearch,anthropic2026webfetch}. We therefore count a cited link as opened if the system successfully fetched it. As search results are encrypted, this measures full-page retrieval only; we cannot
observe how much page content the model receives from search. For DR~Tulu, we count calls to its browsing tool \citep{shao2026dr}. All five systems cite mostly from search results rather than pages they open. Luna and Sol open 9--19\% and 16--30\% of the links they cite (45 of Luna's 65 ResearcherBench reports open no cited page), Sonnet~5 and Opus~5.5 fetch 1.1--3.5\% and 0.2--2.3\%, and DR~Tulu calls its browsing tool only 5 times across all datasets.

\paragraph{QUEST.}
QUEST's visit tool does not return the raw page content to the agent: a separate model (GPT-5-mini) reads  the page and returns an extract conditioned on a goal QUEST states for the visit, consisting of selected evidence and a short summary \citep{xie2026quest}. We count a cited link as opened if QUEST issued a visit to it. Its high open rate therefore does not mean QUEST reads what it cites: the agent sees only extracts filtered toward what it was looking for, which is consistent with its high smoothing despite frequent visits.

\paragraph{Gemini Deep Research.}
The Deep Research API returns only the final report and its citation links, with no browsing or tool steps, so engagement cannot be measured.

\section{Examples of Smoothing in Model Generated Claims}

\subsection{Open-source systems}
\label{sec:open_source_systems}

Examples of generations by open sources systems are provided DR Tulu and Quest are in Table~\ref{tab:opensource-exhibits}.

\begin{table}[t]
\centering
\small
\footnotesize
\setlength{\tabcolsep}{2pt}
\caption{High-severity claims from the open-source systems, with the \textsc{SmoothScore} judge's verdict.}
\begin{tabular}{@{}
  >{\RaggedRight\arraybackslash}p{0.12\textwidth}
  >{\RaggedRight\arraybackslash}p{0.08\textwidth}
  >{\RaggedRight\arraybackslash}p{0.21\textwidth}
  >{\RaggedRight\arraybackslash}p{0.27\textwidth}
  >{\RaggedRight\arraybackslash}p{0.27\textwidth}
@{}}
\toprule
\textbf{Pattern} & \textbf{System} & \textbf{Question}
& \textbf{Claim (score)} & \textbf{Judge verdict} \\
\midrule
Unsupported composition & DR Tulu & \textit{DRB-II ex16:} I am researching the challenges faced by SRAM operating at low voltages, and I hope you can help me survey and compile a technical report on mainstream solutions. The report needs to include the following \ldots &
Alternate Cells: 7T adds explicit precharge--equalize sequence, 8T virtual-ground/zero-precharge, 9T conflict-free write with equalized leakage, 10T quasi-differential read, 11T DDPS for ultra-low leakage, 12T gating and two-stage read. (4) &
The claim over-compresses heterogeneous literature into a fixed one-to-one mapping and misattributes key mechanisms (notably 7T and 11T), while several other pairings are only variant-specific. It requires substantial revision to a qualified, example-based mapping rather than categorical assignments. \\
\addlinespace
Unsupported composition & DR Tulu & \textit{CS2-Expert ex6:} Describe the Sunflower conjecture, progress towards a solution, and related problems. & Decision trees, DNF/branching programs, and communication: Variants of the sunflower conjecture directly constrain the size/width/depth of Boolean formulas and decision trees, with tight results (e.g., $\tilde{\Omega}(k^2)$) for unambiguous DNF; these bounds, proved via explicit constructions and sunflower arguments, carry over to communication protocols and algebraic complexity analogs (e.g., matrix multiplication exponent connections surveyed by Alon). (4) &
The claim over-compresses heterogeneous results: it correctly cites tight unambiguous DNF bounds and connections to communication, and separately, sunflower-variant tools and matrix-multiplication connections, but it incorrectly attributes the tight DNF bound to sunflower arguments, overstates direct constraints on decision trees/branching programs, and overstates transfer to algebraic complexity. It needs substantial narrowing and reattribution. \\
\addlinespace
Premature closure & QUEST & \textit{CS2-Expert ex11:} Survey recent papers on the use of LLMs for 5G network orchestration & Validates that LLM-powered prompt engineering outperforms static ML methods in highly variable O-RAN environments. (5) &
The claim overstates what the evidence shows. The record supports that LLM prompting can beat simple heuristics and be useful in hybrid schemes, but it does not validate outperformance over static ML in highly variable O-RAN environments. \\
\addlinespace
Premature closure & QUEST & \textit{ResearcherBench ex3:} In multimodal pretraining, the current mainstream paradigms are based on image tokens and stable diffusion. Analyzing the latest advancements in these two technical approaches, with reference \ldots & SOTA Across Multiple Benchmarks: DDT-based MLLMs outperform all previously reported models in zero-shot/non-zero-shot evaluation for image-text matching, captioning, question answering (VQA, GQA), object detection, and metric scores (GenEval 0.66, NOCAPs 124.2, EVR L1 14.5, CVS 85.5\%). Even compared to diffusion specialists trained on similar data, DDT shows clear superiority in compositional edits and counterfactual interpolation, owing to its recursive, attribute-rich tokens that mimic natural language structure. (4) &
The claim needs substantial revision: while the quoted metric values are accurate and DDT-based MLLMs are competitive, the evidence shows they do not outperform all previously reported models on several listed benchmarks and provides no support for object detection. \\
\bottomrule
\end{tabular}
\label{tab:opensource-exhibits}
\end{table}

\subsection{Closed-source systems}
\label{sec:closed_source_systems}

Table~\ref{tab:closedsource-exhibits} shows examples of claims generated by closed source systems. 

\begin{table*}[t]
\centering
\small
\footnotesize
\setlength{\tabcolsep}{2pt}
\caption{High-severity claims from the closed-source systems, with the \textsc{SmoothScore} judge's verdict.}
\begin{tabular}{@{}
  >{\RaggedRight\arraybackslash}p{0.12\textwidth}
  >{\RaggedRight\arraybackslash}p{0.08\textwidth}
  >{\RaggedRight\arraybackslash}p{0.21\textwidth}
  >{\RaggedRight\arraybackslash}p{0.27\textwidth}
  >{\RaggedRight\arraybackslash}p{0.27\textwidth}
@{}}
\toprule
\textbf{Pattern} & \textbf{System} & \textbf{Question}
& \textbf{Claim (score)} & \textbf{Judge verdict} \\
\midrule
Scope expansion & Sonnet~5 &
\textit{ResearcherBench ex49:} From a technical perspective, how to understand the similarities and differences between Reinforcement Learning (RL) algorithms and Supervised Fine-Tuning (SFT) in Large Language Models (LLMs), as well as \ldots &
RL, especially when trained with an outcome-based reward, generalizes across both rule-based textual and visual variants, while SFT, in contrast, tends to memorize training data and struggles to generalize out-of-distribution scenarios. (4) &
The claim overstates a mixed and conditional landscape. While there is solid evidence that outcome-based RL can improve OOD generalization in both textual and visual tasks and that SFT often overfits, there is also substantial, direct evidence that SFT can generalize and sometimes surpass RL, and that RL's gains depend on reward design, initialization, and compute. The conclusion needs rebalancing and explicit conditions. \\
\addlinespace
Premature closure & Luna &
\textit{ResearcherBench ex3:} In multimodal pretraining, the current mainstream paradigms are based on image tokens and stable diffusion. Analyzing the latest advancements in these two technical approaches, with reference \ldots &
For multimodal pretraining as a route toward a single general-purpose model, discrete image-token modeling appears more promising than a Stable-Diffusion-centered architecture. It fits naturally with the scaling machinery of language models: images, text, and video can be serialized into one sequence and trained with a common next-token objective. The strongest evidence comes from Emu3, Janus-Pro, Chameleon, and the rapidly improving tokenizers TiTok/FlexTok. (4) &
The claim needs substantial revision. The evidence supports discrete-token, next-token training as a viable and rapidly improving path, but not that it is more promising overall than diffusion-centered approaches; multiple studies show diffusion/flow or hybrid models scaling better and being more compute-efficient, and the `natural fit' is tempered by instability and modality-specific shortcomings. \\
Unsupported composition & Gemini &
\textit{CS2-Expert ex2:} What are some of the methods of detecting GPS location spoofing attack in Connected and Autonomous Vehicles? &
At the mechanical layer, the tight fusion of GNSS data with highly resilient IMU kinematics and LiDAR map-matching ensures that even if a spoofed signal perfectly replicates the cryptographic and spatial signatures of a satellite, it fundamentally cannot alter the physical reality of the road ahead. (4) &
The claim needs substantial revision. The evidence supports that GNSS--IMU--LiDAR/map fusion can greatly harden localization and often detect/mitigate spoofing, but it does not ensure immunity---even against sophisticated, spatially and cryptographically convincing attacks---and has documented failure modes and delays. \\
\bottomrule
\end{tabular}
\label{tab:closedsource-exhibits}
\end{table*}

\end{document}